\documentclass[onecolumn, 11pt, journal]{IEEEtran}

\usepackage[switch]{lineno}
\IEEEoverridecommandlockouts
\usepackage{cite}
\usepackage{amsmath,amssymb,amsfonts}
\usepackage{algorithmic}
\usepackage{float}
\usepackage{subfig}
\usepackage{graphicx}
\usepackage{wrapfig}
\usepackage{enumitem}
\usepackage{multicol}
\graphicspath{
    {img/}
    {img/eval_plots/comp_plots/}
    {img/eval_plots/circles/}{img/eval_plots/pie_charts/}
    {img/segmentation}
}

\usepackage{textcomp}
\usepackage[usenames,dvipsnames]{xcolor}
\usepackage{tikz}
\usetikzlibrary{shapes.geometric, arrows}
\usetikzlibrary{decorations.pathreplacing}
\tikzstyle{arrow} = [thick,->,>=stealth]
\def\BibTeX{{\rm B\kern-.05em{\sc i\kern-.025em b}\kern-.08em
    T\kern-.1667em\lower.7ex\hbox{E}\kern-.125emX}}

\usepackage[pagebackref=true,breaklinks=true,colorlinks,bookmarks=false]{hyperref}
\usepackage[nameinlink]{cleveref}
\Crefname{figure}{Fig.}{Figs.}
\Crefname{equation}{Eq.}{Eqs.}
\usepackage{csquotes}
\usepackage{./tikz/carshapes}
\usepackage{framed}
\renewenvironment{leftbar}[2][\hsize]
{
    \def\FrameCommand
    {
        {\color{#2}\vrule width 3pt}
        \hspace{0pt}
    }
    \MakeFramed{\hsize#1\advance\hsize-\width\FrameRestore}
}
{\endMakeFramed} 

\newcommand{\ie}{i.e.\ }
\newcommand{\eg}{e.g.\ }
\newcommand{\cf}{cf.\ }
\newcommand{\vs}{vs.\ }

\definecolor{slateblue}{rgb}{0.42, 0.35, 0.8}
\definecolor{indianred}{rgb}{0.8, 0.36, 0.36}
\definecolor{mediumseagreen}{rgb}{0.24, 0.7, 0.44}
\definecolor{darkgoldenrod}{rgb}{0.72, 0.53, 0.04}
\definecolor{mediumred-violet}{rgb}{0.73, 0.2, 0.52}
\definecolor{cornflowerblue}{rgb}{0.39, 0.58, 0.93}
\definecolor{goldenyellow}{rgb}{1.0, 0.87, 0.0}
\definecolor{darkseagreen}{rgb}{0.56, 0.74, 0.56}
\definecolor{lightskyblue}{rgb}{0.53, 0.81, 0.98}
\definecolor{mediumslateblue}{rgb}{0.48, 0.41, 0.93}
\definecolor{plum(web)}{rgb}{0.8, 0.6, 0.8}
\definecolor{mediumorchid}{rgb}{0.73, 0.33, 0.83}

\begin{document}

\title{What Should AI See? \\ 
{\LARGE Using the Public's Opinion to Determine the Perception of an AI \vspace{.5em}
}}

\makeatletter
\newcommand{\linebreakand}{%
  \end{@IEEEauthorhalign}
  \hfill\mbox{}\par
  \mbox{}\hfill\begin{@IEEEauthorhalign}
}
\makeatother

\author{\IEEEauthorblockN{\textbf{Robin Chan}$^1$, \textbf{Radin Dardashti}$^2$,  \textbf{Meike Osinski}$^3$, \textbf{Matthias Rottmann}$^1$, \textbf{Dominik Brüggemann}$^4$, \\
\textbf{Cilia Rücker}$^1$, \textbf{Peter Schlicht}$^5$, \textbf{Fabian H{\"{u}}ger}$^6$, \textbf{Nikol Rummel}$^{3,7}$, and \textbf{Hanno Gottschalk}$^1$}\\

\vspace{1.5em}
\IEEEauthorblockA{\normalsize 
1: IZMD, Faculty of Mathematics and Natural Sciences, University of Wuppertal, Germany \\
2: IZWT, School of Humanities, University of Wuppertal, Germany \\
3: Institute of Educational Research, Ruhr-University Bochum, Germany \\
4: IZMD, Chair of Reliability Engineering and Risk Analytics, University of Wuppertal, Germany \\
5: CARIAD SE, Wolfsburg, Germany\\
6: Volkswagen AG, Wolfsburg, Germany \\
7: Center of Advanced Internet Studies, Bochum, Germany
}}

\maketitle

\thispagestyle{empty}
\begin{abstract}
Deep neural networks (DNN) have made impressive progress in the interpretation of image data, so that it is conceivable and to some degree realistic to use them in safety critical applications like automated driving. 
From an ethical standpoint, the AI algorithm should take into account the vulnerability of objects or subjects on the street that ranges from ``not at all'', \eg the road itself, to ``high vulnerability'' of pedestrians. One way to take this into account is to define the cost of confusion of one semantic category with another and use cost-based decision rules for the interpretation of probabilities, which are the output of DNNs.
However, it is an open problem how to define the cost structure, who should be in charge to do that, and thereby define what AI-algorithms will actually ``see''.
As one possible answer, we follow a participatory approach and set up an online survey to ask the public to define the cost structure. We present the survey design and the data acquired along with an evaluation that also distinguishes between perspective (car passenger \vs external traffic participant) and gender. Using simulation based $F$-tests, we find highly significant differences between the groups. These differences have consequences on the reliable detection of pedestrians in a safety critical distance to the self-driving car. We discuss the ethical problems that are related to this approach and also discuss the problems emerging from human--machine interaction through the survey from a psychological point of view. Finally, we include comments from industry leaders in the field of AI safety on the applicability of survey based elements in the design of AI functionalities in automated driving.
\end{abstract}


\section{Introduction}

When human beings and robots interact in public space, robots should take the ethical values of humans into account. As robots are programmed, this requires anticipation of situations where robots could be in conflict with those values and also anticipation of the decision space for such a situation. Ultimately, an algorithmic approach to find and execute a compliant decision is needed. This particularly applies to the case of automated driving, where wrong decisions of the robotic car could be harmful or even deadly to humans. Approaching what has been said above, software engineers encounter various problems:

\begin{enumerate}[label=\bfseries\arabic*), topsep=1ex,itemsep=1ex,] 

    \item In the stage of programming, software engineers might not be aware of the consequences that certain technical settings imply. Ethical problems might just be disregarded. In light of existing \emph{safety by design} procedures in industry, this could be seen as an organizational task to enable software designers to develop \emph{ethics by design} \cite{bundesregierung2018strategie}. 
    
    \item There is no unique ethical system the software developer could refer to. Neither the existing normative theories of ethics need to come to the same decision \cite{allen2005artificial,wallach2008moral,vallor2016technology} nor is it clear, \emph{who} should apply any of these ethical systems: the software engineer, the software company, a philosopher with a PhD degree in ethics, bodies of technical standardization, politicians, judges, or the public? Also the relevance of representation \cite{buolamwini2018gender} and regional variations in ethical decision making \cite{awad2018moral} might play a central role.
    
    \item Whoever makes the decision requires an understanding of the technical matters that determine that decision. What communication strategies could enable \eg citizens to come to a qualified decision?         
    
    \item The perception of a robot and its representation of the world fundamentally differs from human perception. Due to this circumstance, the application of ethical systems developed by and for humans is not straightforward. For instance, the technical perception of a robotic car is usually based on deep neural networks (DNN) that interpret various sensors based on probabilities \cite{Chan2019Dilemma}. 

\end{enumerate}

Concerning this last point, from a technical point of view, DNNs in computer vision output probabilities given high resolution images. Coming to conclusions on the basis of probabilities itself involves ethical decisions, since the consequences of \eg the confusion of a pedestrian with the street could be deadly, whereas the opposite confusion would trigger an unnecessary emergency braking. The perception of the street scene derived from the predicted probabilities therefore changes if the cost of confusion is determined in different ways \cite{Chan18DecisionRules, Chan2019Dilemma}. In particular, different configuration of confusion costs that prioritize the safety of the passengers of the self-driving car over the safety of other external road users may lead to significant changes in perception. 

Just leaving confusion costs aside and just choosing the class with the highest probability using the maximum a posteriori probability principle, also known as the Bayes decision rule, does not avoid the ethical problem, since this amounts to costs of confusion treating any type of confusion the same. Such a cost structure deviates from common ethical intuition. Nevertheless, this ``robotistic'' decision rule is standard in present artificial intelligence (AI) technologies, which gives a good illustration to the first point of disregarded ethical problems mentioned above. In particular, this problem is not explicitly mentioned in any of the recommendations of ethics commissions in autonomous driving on German \cite{difabio2017ethics}, European \cite{bonnefon2020ethics}, or international level \cite{NHTSA2017Automated,zhu2021ethical}.

Elaborating further on \cite{Chan2019Dilemma}, where the ethical problem of decision rules is described and the decision space is presented, in this article we discuss a participatory approach to determine costs of confusions. Between June 2019 and November 2020, participants in our online survey\footnote{\textbf{Visit} \url{http://www-ai.math.uni-wuppertal.de/~chan/poll/} or \url{http://uni-w.de/1hx} for the online survey} entered 5045 proposals as to how severe confusions of random instances adhering to one semantic category (\eg human) with other semantic categories (\eg road) should be judged. The presented instances are extracted from the popular computer vision dataset called Cityscapes \cite{cityscapes}, which is commonly used to train an AI for the perception of street scenes. With our survey, we present an approach to use the public's opinion to determine what an AI system will see. In this article, we experiment with this specific solution strategy to problem point \textbf{2}) and we also position this approach in the field of the ethics of AI, where participatory approaches are recognized as a crucial ingredient for developing trustworthy technologies.

As a further contribution to the second point, we use personal information provided by the survey participants to conduct significance tests for the difference of confusion costs between groups, similar to \cite{awad2018moral}. More precisely, we employ statistical tests to compare the variation within groups to the variation between groups by means of simulation based $F$-tests that are tailored to the data structure of confusion costs. The different confusion cost structures lead to differences in the consequences with regard to human instances in safety critical zones in front of a self-driving car. These are evaluated using software developed in \cite{brueggemann2021}. This allows us to explore the extent to which the judgment of different groups of people and also the ``robotistic'' view can potentially lead to consequences with ethical significance. In general, however, direct statistical assessment of safety of AI systems in automated driving is known to be difficult in practice \cite{gottschalk2021does}.

Concerning point \textbf{3}), we put the online survey in the perspective of current research on human-machine-interaction (HMI) by evaluating the feedback of the survey participants. We also involve industry leaders who give a comment on how our conducted participatory approach can contribute to an actual industrial solution.

Our article does not intend to deliver any solution to the problem of defining confusion costs, \ie to determine the perception of an AI applied in automated driving. In particular, we do not recommend or disrecommend to use the confusion cost matrices obtained from our survey in automated driving. Rather we construct and confront interdisciplinary perspectives on a problem of practical ethics, which emerges from the application of AI in safety critical applications, and investigate potential tensions as well as concordance between those.
It should also be clear that in this work we only investigate one AI perception modality, namely camera based semantic segmentation. Autonomous vehicles, however, are complex systems with various levels of redundancy within and across sensor modalities, including camera, radar, lidar, and ultrasonic detectors. It is beyond the scope of this work to evaluate the consequences of different choices of the cost structure of confusions on the level of such complex systems.

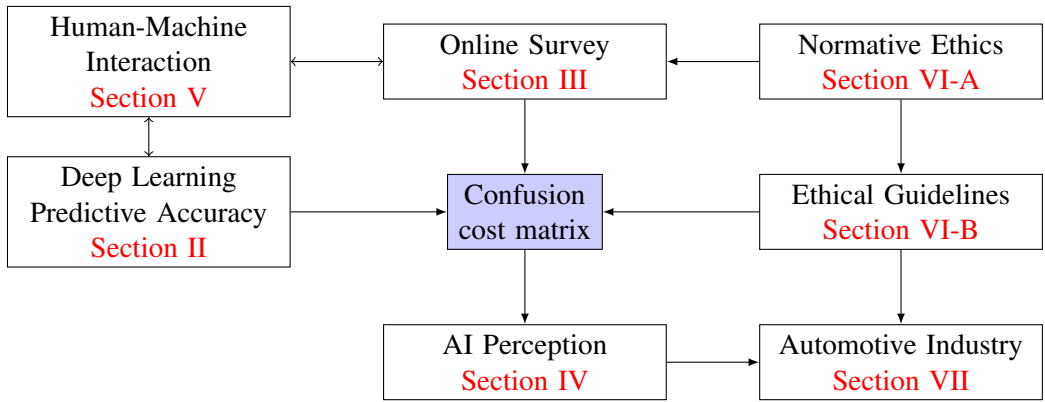
\begin{figure}
    \centering
    \begin{tikzpicture}
\node[rectangle, draw, fill=blue!20, text width=1.8cm, align=center] (cm) at (0,0) {Confusion cost matrix};
\node[draw, text width=3.5cm, align=center] (bayes) at (-5,0) {Deep Learning \\ Predictive Accuracy \\ \Cref{sec:decisionRules}};
\node[draw, text width=3.5cm, align=center] (survey) at (0,2) {Online Survey \\ \Cref{sec:survey}};
\node[draw, text width=3.5cm, align=center] (hmi) at (-5,2) {Human-Machine Interaction \\ \Cref{sec:hmi}};
\node[draw, text width=3.5cm, align=center] (ethics) at (5,2) {Normative Ethics \\ \Cref{sec:normative}};
\node[draw, text width=3.5cm, align=center] (guide) at (5,0) {Ethical Guidelines \\ \Cref{sec:ethical_guidelines}};
\node[draw, text width=3.5cm, align=center] (result) at (0,-2) {AI Perception \\ \Cref{sec:evaluationMethodology}};
\node[draw, text width=3.5cm, align=center] (industry) at (5,-2) {Automotive Industry \\ \Cref{sec:industry}};

\draw[-latex] (bayes) -> (cm);
\draw[-latex] (survey) -> (cm);
\draw[-latex] (guide) -> (cm);
\draw[-latex] (ethics) -- (survey);
\draw[-latex] (ethics) -> (guide);
\draw[-latex, <->] (hmi) -- (survey);
\draw[-latex, <->] (bayes) -- (hmi);
\draw[-latex] (cm) -> (result);
\draw[-latex] (result) -> (industry);
\draw[-latex] (guide) -> (industry);

\end{tikzpicture}
    \caption{The perception of an artificial intelligent (AI) system depends on a \emph{confusion cost matrix}, which is in turn determined by various factors. The sections, in which we will discuss the different factors, are given in this figure.}
    \label{fig:big_picture}
\end{figure}

We provide an overview of the structure of this article in \Cref{fig:big_picture}. This work is organized as follows: In \Cref{sec:decisionRules} we give a mathematical definition of cost-based decision rules and thereby define the space of alternatives. In the subsequent  \Cref{sec:survey} we present our survey-based approach and the survey design. \Cref{sec:evaluationMethodology} presents statistical evidence for the difference between certain groups of people as well as the \emph{robotistic} view, and also evaluates to what extent these differences are safety-relevant. \Cref{sec:hmi} considers the relevant HMI aspects of the survey and explores the difficulties that emerge from just asking the people from the people's perspective.
\Cref{sec:ethics} places this article's approach in the field of ethical discussions on the application of AI, with special emphasis on automated driving. Similarities and Dissimilarities to the infamous trolley problem are discussed and the applicability of existing ethical guidelines are probed.
\Cref{sec:industry} discusses the usefulness and the feasibility of participation based approaches from an industrial point of view and relates such approaches to ethical, legal and technical regulatory frameworks.

\section{Cost-based Decision Rules in Deep Learning for Computer Vision} \label{sec:decisionRules}

The introduction of deep learning, a sub field of machine learning, has enabled advances in many applications of artificial intelligence (AI) that have been considered intractable before, such as computer vision. Computer vision (CV) can be described as the task that deals with enabling machines to gain an high-level understanding of scenes from digital image data. This includes the detection and localization of objects by means of images captured by cameras, which essentially determines what an AI can ``see''. The nowadays established CV pipelines are all based on deep learning and organized around deep neural networks (DNNs). Employing such a type of model contains several gateways by which ethical difficulties may enter. This includes for instance the distribution of object classes in the training data, class weights in the training objectives, or the decision rules incorporated in DNNs to obtain final class predictions. In this section, we elaborate on how ethical problems enter through the latter gateway relating to the concept of decision rules. 

\subsection{Mathematical Formulation of Cost-based Decision Rules and Connection to Standard Decision Principle} \label{sec:decisionRules-math}

\begin{figure}
    \centering
    \input{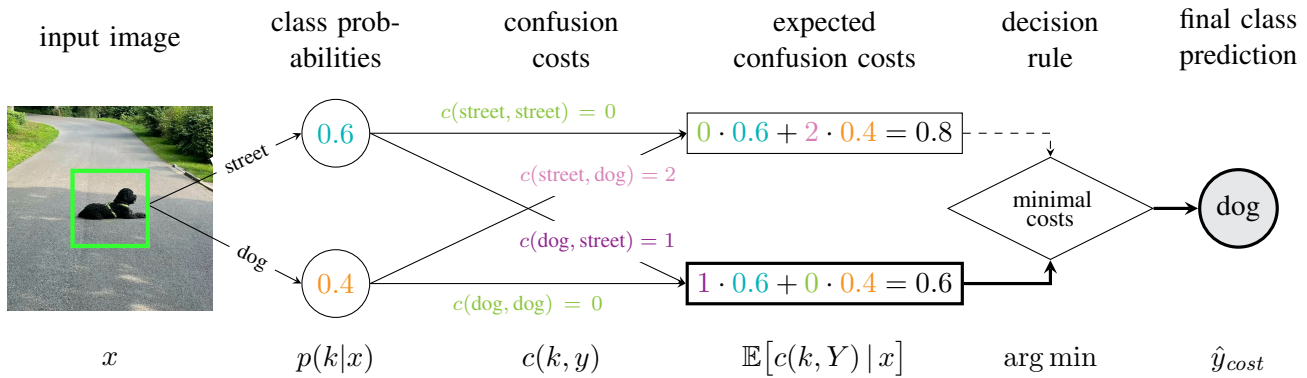}
    \caption{Illustration of a simplified example of a cost-based decision rule in binary classification. Here, the task is to classify the content within the green box in the input image either as \emph{street} (class 1) or \emph{dog} (class 2). In this particular example it is assumed that confusing the class dog with the class street is twice as severe as the other way round, which can be realized by the confusion costs \textcolor{black}{$c(\textrm{street}, \textrm{dog}) = 2$} and \textcolor{black}{$c(\textrm{dog}, \textrm{street}) = 1$}. 
    Then, a cost-based decision rule selects the class that has the lowest expected confusion costs. This results in the class dog as final class prediction although the class probabilities are higher for street (\textcolor{black}{60}\% \vs \textcolor{black}{40}\%). We refer to \Cref{eq:exp-cost-sum} for the general formula for the final class prediction by means of a cost-based decision rule.}
    \label{fig:example_cost_based}
\end{figure}

From a technical point of view, DNNs are typically used as statistical models to estimate a probability that a given input belongs to a certain object category. More formally, let us denote the input by $x$. In computer vision this could \eg be an entire image or even only a single pixel of an image, depending on the type of problem. Further, let us denote the object category corresponding to the input by $y$.
Such a class label is usually encoded with a natural number, \eg $y \in \{1,\ldots,N\}$. Here, $N\in\mathbb{N}$ denotes the number of different classes, which is a design parameter chosen before creating datasets, and thus also before training DNNs to recognize objects in images.
In this regard, we train DNNs to estimate $p(y|x)$, which can be understood as the probability for a given input $x$ having class affiliation $y$. 
Given the probabilistic output of DNNs over all potential classes, the final class prediction is then usually obtained by selecting the class that is assigned the highest probability. This approach is also commonly referred to as \emph{maximum a-posteriori probability} (MAP) principle or \emph{Bayes decision rule} \cite{Fahrmeir1996, Chan18DecisionRules}, which yields the final class prediction
\begin{equation} \label{eq:bayes-decision}
    \hat{y}_\mathit{Bayes}(x) := \mathop{\mathrm{arg\,max}}_{k\in \{1,\ldots,N\}} p(k|x) ~ .
\end{equation}
From decision theory, the Bayes decision rule is known to be merely one example of the more general concept of \emph{cost-based decision rules}. The latter decision principle selects the class that is associated with the lowest expected costs with respect to a classification mistake. To this end, a quantification of the costs of confusion between classes is required.
In more detail, given the cost of confusion $c(k,y) \in \mathbb{R}_{\geq 0}$ between the two classes $k,y \in \{ 1,\ldots, N \}$, the final class prediction for the input $x$ via a cost-based decision rule is then obtained by
\begin{align} \label{eq:exp-cost-sum} 
    \hat{y}_\mathit{cost}(x|c) := \mathop{\mathrm{arg\,min}}_{k\in \{1,\ldots,N\}} \mathbb{E}\big[ c(k,Y) \, | \, x \big]  = \mathop{\mathrm{arg\,min}}_{k\in \{1,\ldots,N\}} \sum_{y=1}^N c(k,y)\, p(y|x) \, , 
\end{align}
where $\mathbb{E}[ c(k,Y) \, | \, x ]$ denotes the expected costs of confusion with respect to class $k$ given input $x$, or, in other words, the expected costs for confusing the considered class $k$ with any other possible class $y\in\{1,\ldots,N\}$. Note that the cost assignments can also be expressed compactly in form of a \emph{confusion cost matrix} $(c(k,y))_{k,y \in \{ 1,\ldots,N \}}$ of size $N \times N$, which will be the subject of discussion in the following sections. For a simplified example illustrating the process of decision making by means of cost-based decision rules, we refer to \Cref{fig:example_cost_based}. At this point, we want to emphasize that the final class prediction via cost-based decision rules is not affected by the absolute values but by the relative differences of confusion costs. More formally, the cost-based decision remains unaffected by a constant non-negative factor $\lambda \in \mathbb{R}_{\geq 0}$ applied to the confusion costs, \ie
\begin{align} \label{eq:cost-absolute} 
    \hat{y}_\mathit{cost}(x|c) 
    = \mathop{\mathrm{arg\,min}}_{k\in \{1,\ldots,N\}} \mathbb{E}\big[ c(k,Y) \, | \, x \big] 
    = \mathop{\mathrm{arg\,min}}_{k\in \{1,\ldots,N\}} \lambda\, \mathbb{E}\big[ c(k,Y) \, | \, x \big]
    = \mathop{\mathrm{arg\,min}}_{k\in \{1,\ldots,N\}} \mathbb{E}\big[ \lambda\, c(k,Y) \, | \, x \big] ~.
\end{align}

Still, a question that naturally arises in this context is how to choose the quantities $c(k,y) ~\forall~k,y \in \{ 1,\ldots, N \}$. For example in scenarios of automated driving, for the two classes ``street'' and ``human'', how should the cost of the confusion be valuated? Moreover, should the cost of confusion between these two classes be symmetric, \eg is overlooking humans in favor of the street as severe as the other way round? Common human intuition would suggest that confusion costs should be different depending on the type of confusion. However, it remains an open question, what values should explicitly be used.

As a matter of fact, however, the confusion costs are already implicitly defined in DNNs when employing the standard Bayes decision rule. Returning now to the mentioned statement that the Bayes decision rule is merely one example of a cost-based decision rule, it turns out that this \emph{standard decision principle incorporates constant confusion costs, weighting each type of confusion equally serious}. More precisely, for each type of confusion between two classes $k$ and $y$, the Bayes decision is based on the cost valuation
\begin{align} \label{eq:simple-symm-cost}
    c_\mathit{robot}(k,y) = 
    \begin{cases}
    0 & \mathrm{if~} k=y \\
    1 & \mathrm{if~} k\neq y
    \end{cases} 
    \quad \forall ~ k,y \in \{ 1,\ldots, N \} ~,
\end{align}
resulting in the connection between Bayes decision rule and the latter cost function
\begin{align} \label{eq:cost-bayes}
    \begin{split}
    \hat{y}_\mathit{Bayes}(x) 
    \stackrel{(\ref{eq:bayes-decision})}{=}& \mathop{\mathrm{arg\,max}}_{k\in \{1,\ldots,N\}} p(k|x) 
    \stackrel{(\ref{eq:simple-symm-cost})}{=} \mathop{\mathrm{arg\,min}}_{k\in \{1,\ldots,N\}} \mathbb{E}\big[ c_\mathit{robot}(k,Y) \, | \, x \big]
    \stackrel{(\ref{eq:exp-cost-sum})}{=} \hat{y}_\mathit{cost}(x|c_\mathit{robot}) ~,
    \end{split}
\end{align}
\cf also \cite{Fahrmeir1996, Chan2019Dilemma}. In this way, the Bayes decision principle aims at \emph{minimizing the chance of any incorrect prediction}, since there is no distinction in the type of error according to \Cref{eq:simple-symm-cost}, which is equivalent to \emph{maximizing the model's predictive accuracy}. We refer to this kind of decision making as the \emph{robotistic} attitude. Furthermore, as revealed in \cite{Chan2019Dilemma}, the Bayes decision rule leads to the ethical dilemma of machine learning models in general that, on the one hand, it is not evident what confusion cost values should explicitly be used, but on the other hand, confusion cost values in conflict with common human intuition are already set implicitly by default. In this work, we study different confusion cost valuations in the context of street scenes, with the costs determined by the public in an online survey. While doing so, we study qualitative differences in the obtained cost matrices for different groups of survey participants.

\subsection{Values and Tradeoffs in Cost-based Decision Rules} \label{sec:Tradeoffs}

Value judgments are ubiquitous in science and technology and have been a main issue in discussions on the ethics and philosophy of science and technology (\eg \cite{douglas2009science,longino2018fate,brown2020science}). They can enter at various steps of the inquiry and development when different epistemic and non-epistemic goals are in conflict, thereby leading to tradeoff situations. A specific choice within this tradeoff space amounts to a possible entry point for values. Quite frequently, these choices remain implicit or are not even recognized as ``choices'' and their corresponding tradeoff space remains similarly unknown. These contingencies of the process of inquiry should in morally significant situations be the subject of reflective evaluation \cite{brown2020science}.

These tradeoff situations have already been recognized in the work of software engineers that use machine learning techniques, see \eg \cite{biddle2020predicting}. Issues regarding the opacity of the AI system and the potential biases present in the dataset from which the algorithm learns, have been central topics of discussion in the ethics of AI literature \cite{sep-ethics-ai}. Once these potentially ethical issues in the choice of these tradeoff situations have been recognized, a plethora of policy recommendations and ethical principles for more transparency and explainability in the design process have been suggested, see \eg the ethics guidelines of AI High-level expert group of the European Commission \cite{aihleg2019} or for an overview \cite{jobin2019global}. As was shown in \cite{jobin2019global} there seems to be a global convergence regarding the principles that should underlie the ethical use of AI, while a divergence was recognized regarding specific implementations of these. Thus, a detailed ethical analysis complementing available ethical guidelines is essential as it is explicitly requested in \cite[p. 23]{bonnefon2020ethics}.

In \Cref{sec:decisionRules-math} the possible choices for the parameters in the costs of confusion in \Cref{eq:exp-cost-sum} provides the tradeoff space within which various epistemic and non-epistemic values need to be assessed and possibly weighed against each other. This can easily be illustrated by a comparison between the recall and the precision in the case of human classification, \cf also \Cref{fig:rec-prc}.
By varying the costs of confusion, one may maximize the recall (or equivalently the sensitivity) to the point where there are no false negatives. This, however, goes hand in hand with a decrease in the precision, as the number of false positives similarly increases with the variation of cost values. This is a generic feature of any non-perfect but realistic classifier. 
So one may be able to identify all humans correctly, but only at the cost of identifying additionally many non-human things as humans. This would lead to a significant decrease in the practicability of autonomous vehicles, as it would mistakenly hit the brakes too frequently. Here practicability enters as a value, which if not satisfied would significantly diminish the usability of any autonomous vehicle. 

In this particular context, an ethical assessment of the outlined tradeoff is non-trivial as there are utilitarian reasons for the introduction of autonomous vehicles which would warrant an increase in practicability (\cf \cite{difabio2017ethics}, although these are critically discussed \cite{lin2013ethics, hevelke2015responsibility}), thereby making practicability a requirement, which nevertheless may be in tension with the individual lives that are put at risk by decreasing the recall.

This tradeoff between the recall and precision, or, the identification of humans and the practicability of the autonomous vehicle, respectively, illustrates just one of many values that may impact, if made explicit, the determination of the confusion cost matrix. Other examples would be fuel efficiency and speed but also issues of privacy may impact the tradeoff space. In this paper we are particularly concerned with the tradeoff between accuracy, as implemented by the Bayes decision rule, \cf \Cref{eq:cost-bayes}, and the public's opinion, as implemented in a survey-based approach in determining the cost confusion matrix. We will further elaborate on the tradeoff and consider the guidelines provided by the ethics commission \cite{difabio2017ethics}. 
\begin{figure}
    \centering
    \subfloat[computation of the classification performance metrics recall and precision]{\scalebox{0.8}{
    \begin{tikzpicture}
    
    \def\xr{-3.6}
    \def\xp{3.2}

    \node at (\xr,0) {$\mathrm{recall} = $};
    \fill [orange!40] (\xr+1.5,1.66) rectangle (\xr+3,1);
    \draw [color=green!80, ultra thick] (\xr+1.5,1.66) rectangle (\xr+4.5,0.33);
    \draw [color=blue!80, ultra thick] (\xr+1,2) rectangle (\xr+3,1);
    \draw (\xr+1,0) -- (\xr+4.5,0);
    \fill [orange!40] (\xr+1,-0.5) rectangle (\xr+3,-1.5);
    \draw [color=green!80, ultra thick] (\xr+1.5,-0.84) rectangle (\xr+4.5,-2.17);
    \draw [color=blue!80, ultra thick] (\xr+1,-0.5) rectangle (\xr+3,-1.5);
    \node at (\xr+2.0,2.25) {\footnotesize \textcolor{blue!80}{target}};
    \node at (\xr+3.8,1.9) {\footnotesize \textcolor{green!60}{prediction}};
    \node at (\xr+2.0,-.25) {\footnotesize \textcolor{blue!80}{target}};
    \node at (\xr+3.8,-.6) {\footnotesize \textcolor{green!60}{prediction}};
    
    
    \node at (\xp,0) {$\mathrm{precision} = $};
    \fill [orange!40] (\xp+1.7,1.66) rectangle (\xp+3.2,1);
    \draw [color=green!80, ultra thick] (\xp+1.7,1.66) rectangle (\xp+4.7,0.33);
    \draw [color=blue!80, ultra thick] (\xp+1.2,2) rectangle (\xp+3.2,1);
    \draw (\xp+1.2,0) -- (\xp+4.7,0);
    \fill [orange!40] (\xp+1.7,-0.84) rectangle (\xp+4.7,-2.17);
    \draw [color=green!80, ultra thick] (\xp+1.7,-0.84) rectangle (\xp+4.7,-2.17);
    \draw [color=blue!80, ultra thick] (\xp+1.2,-0.5) rectangle (\xp+3.2,-1.5);
    \node at (\xp+2.2,2.25) {\footnotesize \textcolor{blue!80}{target}};
    \node at (\xp+4,1.9) {\footnotesize \textcolor{green!60}{prediction}};
    \node at (\xp+2.2,-.25) {\footnotesize \textcolor{blue!80}{target}};
    \node at (\xp+4,-.6) {\footnotesize \textcolor{green!60}{prediction}};
    
    
\end{tikzpicture}
    }}
    \qquad\qquad
    \subfloat[tradeoff between recall and precision]{\scalebox{0.8}{
    \qquad\begin{tikzpicture}
    
    \def\x{0}
    \def\y{1.75}
    \def\z{3.5}

    
    \draw [color=blue!80, ultra thick] (\x,\z) rectangle (\x+1,\z+1);
    \draw [color=green!80, ultra thick] (\x+.5,\z) rectangle (\x+1,\z+.5);
    \node at (\x+4.2,\z+.7) {$\mathrm{recall} = 0.25 = 25\%$};
    \node at (\x+4.08,\z+.2) {$\mathrm{precision} = \mathbf{1.00} = \mathbf{100}\%$};
    
    \draw [color=blue!80, ultra thick] (\x,\y) rectangle (\x+1,\y+1);
    \draw [color=green!80, ultra thick] (\x+.293,\y-.293) rectangle (\x+1.293,\y+.707);
    \draw [color=gray!75, densely dashed] (\x+.5,\y) rectangle (\x+1,\y+.5);
    \node at (\x+4.2,\y+.7) {$\mathrm{recall} = 0.50 = 50\%$};
    \node at (\x+3.9,\y+.2) {$\mathrm{precision} = 0.50 = 50\%$};
    
    \draw [color=blue!80, ultra thick] (\x,0) rectangle (\x+1,1);
    \draw [color=green!80, ultra thick] (\x,1) rectangle (\x+1.5,-0.5);
    \draw [color=gray!75, densely dashed] (\x+.293,-.293) rectangle (\x+1.293,.707);
    \draw [color=gray!50, densely dashed] (\x+.5,0) rectangle (\x+1,.5);
    \node at (\x+4.38,.5) {$\mathrm{recall} = \mathbf{1.00} = \mathbf{100}\%$};
    \node at (\x+3.9,0) {$\mathrm{precision} = 0.44 = 44\%$};
    
\end{tikzpicture}\quad
    }}
    \vspace{.1cm}
    \caption{Illustration of the classification performance metrics recall and precision.
    (a) The recall is the fraction of the amount of overlap between target and prediction divided by the amount of the target, while the precision is the fraction of the amount of overlap between target and prediction divided by the amount of the prediction.
    (b) The sensitivity of predictions can be adjusted by varying thresholds (or implicitly by varying confusion costs), which subsequently increases the amount of predictions. In this way the recall can be maximized, however to the detriment of decreasing the precision, illustrating the tradeoff between these two classification performance metrics. 
    }
    \label{fig:rec-prc}
\end{figure}
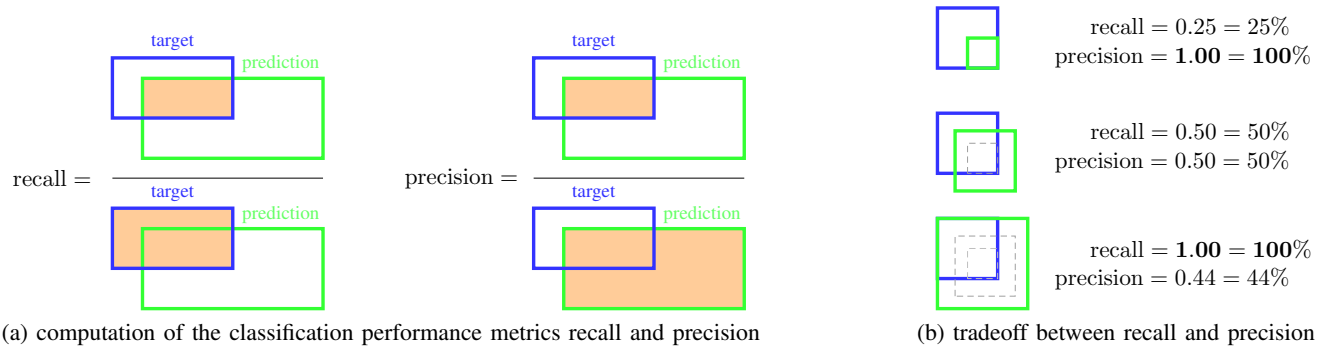

\section{Survey Design for the Valuation of Confusion Costs in Street Scenes} \label{sec:survey}
In the context of automated driving, cameras are one of the main components for perceiving the environment of a self-driving car. To this end, AI algorithms are deployed in order to interpret the captured images. They classify objects into predefined classes based on estimated probabilities that an object belongs to a certain class. As described in the previous \Cref{sec:decisionRules}, the AI is usually programmed to select the class that has the highest probability \cite{Fahrmeir1996, Mehryar12mlfoundation, Chan18DecisionRules}, see also \Cref{eq:bayes-decision}. Hence, all types of confusion of classes are implicitly treated to be equally serious, \cf \Cref{eq:cost-bayes}. However, in terms of safety this decision principle with constant costs of confusion is not necessarily optimal and ethically differs from common human intuition. 

For instance, assume the case in which the AI suggests the class ``road'' and the class ``human'' for an object of interest with an estimated probability of 51\% and 49\%, respectively. In this particular case, strictly trusting the AI and choosing \emph{road over human} because of the slightly higher probability could lead to fatal consequences. From a safety point of view, it therefore seems recommendable that confusions are assessed according to their type as well as the occurrence of possible dangers.

In this section, we present the conducted survey, in which we ask the survey participants to provide confusion costs. We aim at examining the ethical attitude of the public in regard of safety-critical confusions in street scenes. Ultimately, the submitted values should help the AI to adapt its perception to human ethical intuition. 
Starting now with the design of the survey, we distinguish between two possible perspectives to view street scenarios from, either
\begin{center}
        \begin{minipage}{.7\textwidth}
        \begin{multicols}{2}
        \begin{itemize}
            \item[1)] \emph{passenger of the self-driving car},
            \item[2)] or \emph{external traffic participant}.
        \end{itemize}
        \end{multicols}
    \end{minipage}
\end{center}
\vspace{6pt}
We focus on these two perspectives as putting passenger first \vs putting other road users first has already been subject of intense public debate \cite{egocar}. One of these two perspectives is randomly assigned to each participant at the start of the survey and last until the survey is stopped. Based on the perspective, potential confusions in different street scenes are evaluated according to the participants' subjective sense of the severity of consequences.
The street scenes that are shown during the survey are extracted from the publicly available Cityscapes dataset \cite{cityscapes}. Cityscapes is a large collection of images containing diverse images recorded in urban street scenarios in 50 different European cities and it is a popular computer vision benchmark to evaluate how well deep neural networks perform at interpreting complex traffic scenes. There are 19 object classes available in this dataset to be classified. For the sake of simplicity, we reduce the number of classes in the survey by considering class aggregates. We distinguish between six (more superficial) object categories, which are namely 1) driveable, 2) nondriveable, 3) static, 4) info, 5) human, and 6) dynamic, see also \Cref{fig:mapping} for an overview.

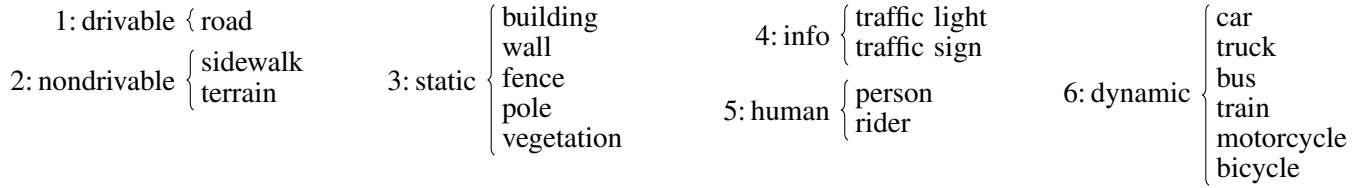
\begin{figure}
    \centering
    \begin{tikzpicture}

\def\aw{0}
\def\bw{0}
\def\cw{4}
\def\dw{7.5}
\def\ew{7.5}
\def\fw{12.5}

\def\ah{0}
\def\bh{-0.6}
\def\ch{0}
\def\dh{0}
\def\eh{-1.0}
\def\fh{0}

\node[text width=4cm] at (\aw,\ah) {road};
\draw[decorate, decoration={brace,mirror}]  ({\aw-2.1},{\ah+0.15}) -- node[left=0.6ex] {{1:\,drivable}}  ({\aw-2.1},{\ah-0.15});

\node[text width=4cm, anchor=mid] at (\bw,\bh) {sidewalk};
\node[text width=4cm, anchor=mid] at (\bw,{\bh-0.4}) {terrain};
\draw[decorate, decoration={brace,mirror}]  ({\bw-2.1},{\bh+0.2}) -- node[left=0.6ex] {{2:\,nondrivable}}  ({\bw-2.1},{\bh-0.6});

\node[text width=4cm, anchor=mid] at (\cw,\ch) {building};
\node[text width=4cm, anchor=mid] at (\cw,{\ch-0.4}) {wall};
\node[text width=4cm, anchor=mid] at (\cw,{\ch-0.8}) {fence};
\node[text width=4cm, anchor=mid] at (\cw,{\ch-1.2}) {pole};
\node[text width=4cm, anchor=mid] at (\cw,{\ch-1.6}) {vegetation};
\draw[decorate, decoration={brace,mirror}]  ({\cw-2.1},{\ch+0.2}) -- node[left=0.6ex] {{3:\,static}}  ({\cw-2.1},{\ch-1.8});

\node[text width=1.9cm, anchor=mid] at (\dw+0.15,\dh) {traffic light};
\node[text width=1.9cm, anchor=mid] at (\dw+0.15,{\dh-0.4}) {traffic sign};
\draw[decorate, decoration={brace,mirror}]  ({\dw-0.9},{\dh+0.2}) -- node[left=0.6ex] {{4:\,info}}  ({\dw-0.9},{\dh-0.6});

\node[text width=1.6cm, anchor=mid] at (\ew,\eh) {person};
\node[text width=1.6cm, anchor=mid] at (\ew,{\eh-0.4}) {rider};
\draw[decorate, decoration={brace,mirror}]  ({\ew-0.9},{\eh+0.2}) -- node[left=0.6ex] {{5:\,human}}  ({\ew-0.9},{\eh-0.6});

\node[text width=2.0cm, anchor=mid] at (\fw,\fh) {car};
\node[text width=2.0cm, anchor=mid] at (\fw,{\fh-0.4}) {truck};
\node[text width=2.0cm, anchor=mid] at (\fw,{\fh-0.8}) {bus};
\node[text width=2.0cm, anchor=mid] at (\fw,{\fh-1.2}) {train};
\node[text width=2.0cm, anchor=mid] at (\fw,{\fh-1.6}) {motorcycle};
\node[text width=2.0cm, anchor=mid] at (\fw,{\fh-2.0}) {bicycle};
\draw[decorate, decoration={brace,mirror}]  ({\fw-1.1},{\fh+0.2}) -- node[left=0.6ex] {{6:\,dynamic}}  ({\fw-1.1},{\fh-2.2});

\end{tikzpicture}
    \vspace{-.5cm}
    \caption{Class aggregates of Cityscapes object classes that are used in the survey. Note that the class \emph{sky} is ignored in the survey since the non-detection of the sky does not cause hazardous street scenarios.}
    \label{fig:mapping}
\end{figure}

\begin{figure}
  \begin{minipage}{0.475\linewidth}
      \begin{figure}[H]
        \centering
         \includegraphics[height=3.7cm]{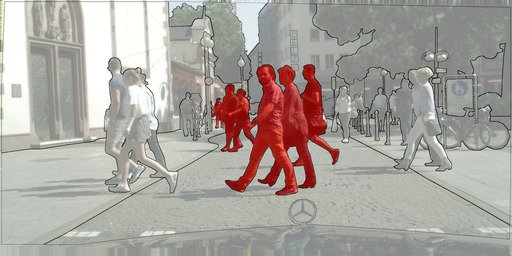}\\
         \vspace{.5cm}
         \caption{An example of a traffic scene that the survey participants evaluate. Marked with red is the object class, in this case human, for which a possible confusion is assessed.}\label{fig:traffic_scene_example}
      \end{figure}
  \end{minipage}%
  \hspace{0.05\linewidth}%
  \begin{minipage}{0.475\linewidth}
      \begin{figure}[H]
        \centering
         \includegraphics[height=3.7cm]{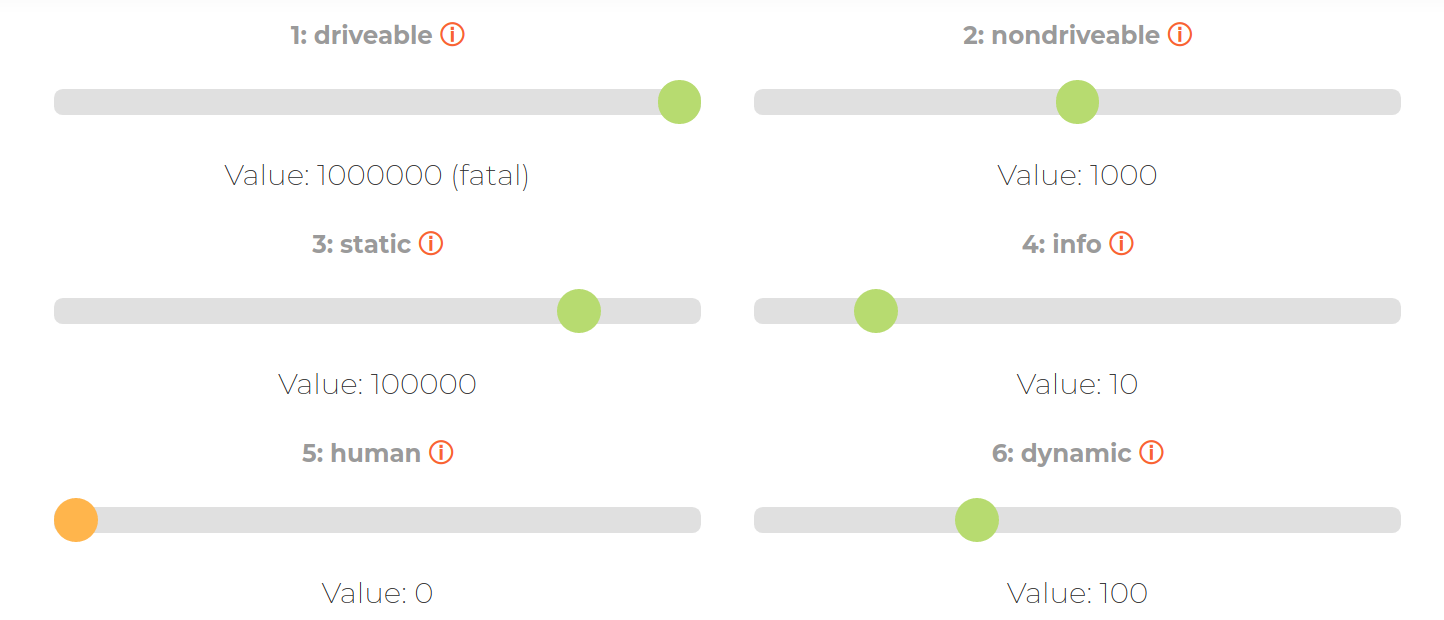}\\
         \vspace{.5cm}
         \caption{The survey participants assess confusions in the displayed street scene via sliders with confusion cost values $1=10^0, 10=10^1$ up to $1\text{M}=10^6$. For the correct class label the cost value is fixed at 0.} \label{fig:slider}
      \end{figure}
  \end{minipage}
\end{figure}

In each of the street scene images displayed in the survey, one instance corresponding to exactly one of the six outlined classes is highlighted, see \Cref{fig:traffic_scene_example} for an example. Given this displayed image, the task of the survey participants is to assess a potential confusion with an instance from another class, see \Cref{fig:slider} for an illustration of the interface from the survey. With a confusion between car and bus (a confusion between two objects within the same object category) being a reference mistake, the survey participants should assess how severe it is if the highlighted instance is incorrectly assigned to one of the remaining five classes. In other words, the \emph{severity of overlooking} the highlighted object must be valuated. One can choose out of 10 (\emph{nearly harmless}), 100 (\emph{fairly harmless}) up to 1M (\emph{fatal}) times more severe than the reference mistake, or 1 (\emph{marginal}) if the confusion is as serious as the reference mistake. Thus, in total there are seven available levels of severity to choose from. With respect to the same class, naturally no confusion costs should incur, therefore the cost value is fixed to 0. Note that the exponential scale of confusion costs is due to the particular architecture of deep neural networks, which have shown gradual changes in their outputs only with confusion costs of this exponential magnitude, \cf \cite{Chan2019Dilemma}.

For a more in-depth analysis of the survey data, the participants are additionally asked to voluntarily provide the following personal information on:
\begin{center}
        \begin{minipage}{.7\textwidth}
        \begin{multicols}{3}
        \begin{itemize}
            \item gender,
            \item age,
            \item graduation,
            \item field of work/study,
            \item whether they own a driver's license,
            \item and what they use as main transportation.
        \end{itemize}
        \end{multicols}
    \end{minipage}
\end{center}
\vspace{6pt}
Besides the perspective, such meta data can also be used in order to form groups and analyze the collected survey data for statistical relationships, which will be subject of discussion in what follows.

\section{Evaluation Methodology for Confusion Costs and Numerical Results} \label{sec:evaluationMethodology}

At the time of writing 520 people have participated in the survey, answering 5,045 questions in total. Given this collected data, we inspect for differences in the confusion costs valuation in this section.
To this end, we form the groups to investigate by restricting the survey data by means of the assigned perspective and based on the gender of the survey participants. These two characteristics form data subsets of roughly the same size:
\begin{center}
    \begin{minipage}{.75\textwidth}
        \begin{multicols}{2}
        \begin{itemize}
            \item \emph{passenger}: 2744 answers,
            \item \emph{external}: 2301 answers,
            \item \emph{female}: 2444 answers,
            \item \emph{male}: 2523 answers,
        \end{itemize}
        \end{multicols}
    \end{minipage}
\end{center}
\vspace{6pt}
besides being sufficiently large to compare them statistically. Note that the groups \emph{passenger} and \emph{external} denote the perspectives corresponding to a passenger of the self-driving car and an external traffic participant, respectively.

\subsection{Comparison of Confusion Cost Values provided by the Survey Data} \label{sec: results-matrices}

\begin{figure*}
    \centering
    \scalebox{0.875}{
    \begin{tikzpicture}

\def\w{-14.3}
\def\wp{-12.1}
\def\gapx{1.2}
\def\gapy{0.472}


\node [align=left,rotate=45,text width=1.65cm] at ({\wp+0*\gapx},2.2) {``drivable''};
\node [align=left,rotate=45,text width=1.65cm] at ({\wp+1*\gapx},2.2) {``nondrivable''};
\node [align=left,rotate=45,text width=1.65cm] at ({\wp+2*\gapx},2.2) {``static''};
\node [align=left,rotate=45,text width=1.65cm] at ({\wp+3*\gapx},2.2) {``info''};
\node [align=left,rotate=45,text width=1.65cm] at ({\wp+4*\gapx},2.2) {``human''};
\node [align=left,rotate=45,text width=1.85cm] at ({\wp+5*\gapx},2.2) {``dynamic''};

\node [align=left,text width=1.65cm, anchor=mid] at ({\wp+6*\gapx},1.2-0*\gapy) {``drive''};
\node [align=left,text width=1.65cm, anchor=mid] at ({\wp+6*\gapx},1.2-1*\gapy) {``nondrive''};
\node [align=left,text width=1.65cm, anchor=mid] at ({\wp+6*\gapx},1.2-2*\gapy) {``static''};
\node [align=left,text width=1.65cm, anchor=mid] at ({\wp+6*\gapx},1.2-3*\gapy) {``info''};
\node [align=left,text width=1.65cm, anchor=mid] at ({\wp+6*\gapx},1.2-4*\gapy) {``human''};
\node [align=left,text width=1.65cm, anchor=mid] at ({\wp+6*\gapx},1.2-5*\gapy) {``dynamic''};

\node [align=center] at (\w,0) {$\bar{C}_P=$};
\node [align=center] at ({\w+4.5},0) 
{$\begin{pmatrix}
   0 & 10^{4.12} & 10^{3.97} & 10^{3.75} & 10^{4.74} & 10^{3.97} \\
10^{3.90} &    0 & 10^{2.70} & 10^{3.18} & 10^{3.42} & 10^{3.30} \\
10^{3.34} & 10^{2.50} &    0 & 10^{2.70} & 10^{3.51} & 10^{3.07} \\
10^{2.96} & 10^{2.96} & 10^{2.76} &    0 & 10^{3.51} & 10^{3.13} \\
10^{3.72} & 10^{3.00} & 10^{3.05} & 10^{3.41} &    0 & 10^{3.18} \\
10^{3.41} & 10^{3.17} & 10^{3.05} & 10^{3.41} & 10^{3.18} &    0 \\
\end{pmatrix}$};

\node [align=center] at (-3.8,0) {$\bar{C}_E=$};
\node [align=center] at (0.7,0) 
{$\begin{pmatrix}
   0 & 10^{4.42} & 10^{4.36} & 10^{4.00} & 10^{5.51} & 10^{4.71} \\
10^{3.71} &    0 & 10^{2.72} & 10^{3.06} & 10^{3.99} & 10^{3.40} \\
10^{3.70} & 10^{2.13} &    0 & 10^{2.41} & 10^{3.56} & 10^{3.46} \\
10^{2.97} & 10^{2.46} & 10^{2.79} &    0 & 10^{3.70} & 10^{3.08} \\
10^{4.03} & 10^{3.04} & 10^{3.16} & 10^{3.40} &    0 & 10^{3.50} \\
10^{3.77} & 10^{2.84} & 10^{3.14} & 10^{3.34} & 10^{3.14} &    0 \\
\end{pmatrix}$};

\draw[decorate, decoration={brace}]  (-2.9,1.6) -- node[above=0.4ex] {\textit{true target class label in columns}} (4.2,1.6);
\draw[decorate, decoration={brace}]  (4.8,1.5) -- node[below=0.6ex, rotate=90] {\textit{prediction in rows}}  (4.8,-1.5);

\node [align=center] at (\w,-3) {$\bar{C}_F=$};
\node [align=center] at ({\w+4.5},-3) 
{$\begin{pmatrix}
   0 & 10^{4.45} & 10^{4.25} & 10^{4.23} & 10^{5.65} & 10^{4.47} \\
10^{4.03} &    0 & 10^{2.72} & 10^{3.32} & 10^{4.06} & 10^{3.79} \\
10^{3.83} & 10^{2.26} &    0 & 10^{2.48} & 10^{3.65} & 10^{3.36} \\
10^{3.29} & 10^{2.92} & 10^{2.81} &    0 & 10^{3.84} & 10^{3.41} \\
10^{4.14} & 10^{3.09} & 10^{3.28} & 10^{3.54} &    0 & 10^{3.44} \\
10^{4.03} & 10^{3.07} & 10^{3.23} & 10^{3.56} & 10^{3.22} &    0 \\
\end{pmatrix}$};

\node [align=center] at (-3.8,-3) {$\bar{C}_M=$};
\node [align=center] at (0.7,-3) 
{$\begin{pmatrix}
  0 & 10^{4.33} & 10^{4.03} & 10^{3.58} & 10^{4.60} & 10^{4.17} \\
10^{3.74} &    0 & 10^{2.76} & 10^{2.98} & 10^{3.39} & 10^{3.04} \\
10^{3.33} & 10^{2.43} &    0 & 10^{2.64} & 10^{3.48} & 10^{3.15} \\
10^{2.74} & 10^{2.54} & 10^{2.76} &    0 & 10^{3.39} & 10^{2.88} \\
10^{3.73} & 10^{3.02} & 10^{3.04} & 10^{3.32} &    0 & 10^{3.24} \\
10^{3.30} & 10^{2.98} & 10^{3.03} & 10^{3.24} & 10^{3.17} &    0 \\
\end{pmatrix}$};


\end{tikzpicture}
    }
    \caption{Average confusion cost matrices determined by different groups formed by the assigned perspective and gender of the survey participants, namely these groups are \emph{passenger} ($\bar{C}_P$), \emph{external} ($\bar{C}_E$), \emph{female} ($\bar{C}_F$), and \emph{male} ($\bar{C}_M$), \cf also \Cref{sec:evaluationMethodology}. The matrices are read as follows: \eg for the fifth entry in the first row, how fatal is the confusion if the model predicts the class \emph{driveable} but the actual true class label is \emph{human}?}
    \label{fig:average-cost-matrices}
\end{figure*}
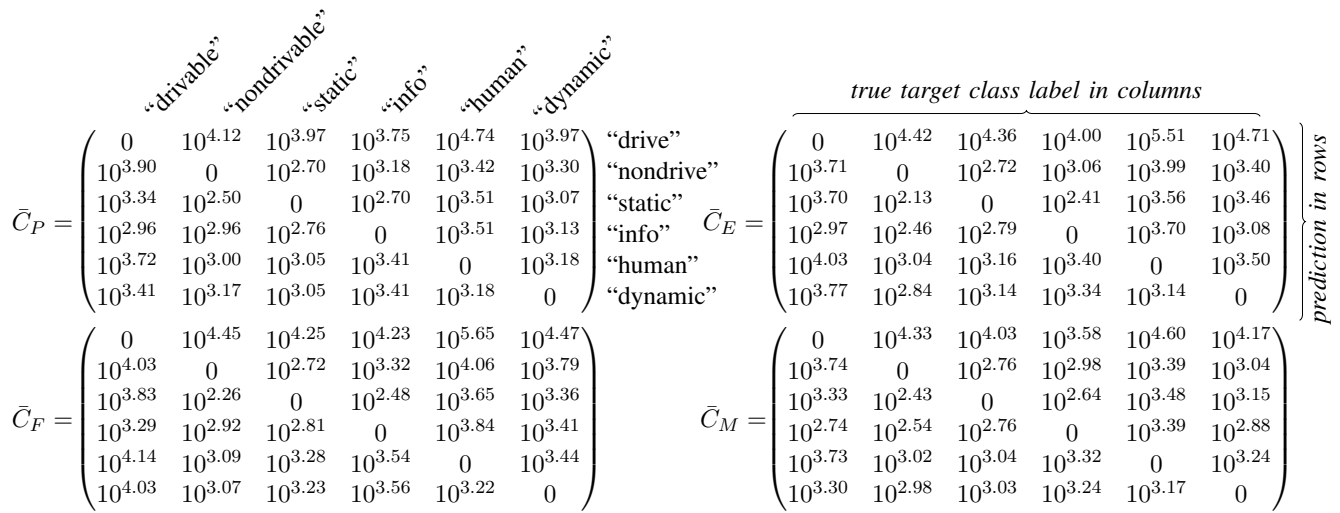

The average confusion cost values of the outlined groups are displayed in the form of cost matrices in \Cref{fig:average-cost-matrices}. Comparing the matrices provided by the assigned perspective, \ie \emph{passenger of the self-driving car \vs external traffic participant}, we observe that the cost valuations are in a similar range. Some noteworthy exceptions are related to predictions of the class ``drivable''. More precisely, the confusion of either ``human'' or ``dynamic'' in favor of ``drivable'' is valuated considerably more severe by the group \emph{external} (costs of $10^{5.51}$ and $10^{4.71}$ \vs $10^{4.74}$ and $10^{3.97}$, respectively). Focusing now only on the class ``human'', confusions involving this vulnerable class are generally assigned higher costs by the group \emph{external} as well. The corresponding column in the respective matrices (which represents the submitted cost values when a human instance is highlighted in the image displayed in the survey) let us conclude that by means of the submitted values, the group \emph{external} tends towards a more human sensitive cost assignment than the group \emph{passenger}.

Similar observations can be made when comparing the cost matrices provided by gender, \ie \emph{female} \vs \emph{male}. Here, the cost values of the group \emph{female} show to be higher in general for most types of confusion. In light of the absolute values, this is most significant for the confusion ``driveable'' as prediction and ``human'' as true target class label with cost values $10^{5.65}$ and $10^{4.60}$, respectively. Focusing again only on the class ``human'', confusions involving this vulnerable class are also assigned clearly higher costs by the group \emph{female}. In fact, with respect to the class ``human'', none of the cost valuations provided by the group \emph{male} exceed those of the group \emph{female}. This let us conclude that by means of the submitted values, the group \emph{female} tends towards a more human sensitive cost assignment than the group \emph{male}.

This leads to another finding that is related to the tradeoff between recall and precision. As already introduced in \Cref{sec:Tradeoffs}, recall can be maximized by increasing the prediction sensitivity, but possibly sacrificing precision. In light of confusion cost matrices, the prediction sensitivity with respect to one class can be varied by the cost values within the corresponding column. This consequently counteracts with the cost values in the row of the considered class in the confusion matrix, which in turn controls the predictive precision. We have already concluded that both the groups \emph{external} and \emph{female} show to have a more human sensitive cost assignment than the groups \emph{passenger} and \emph{male}, respectively. This might be still valid given the submitted cost values and the design of the survey, in which the participants explicitly provided cost values for the severity of overlooking humans, \cf \Cref{sec:survey}. 

However, from a technical point of view, this does not necessarily imply better recall with respect to human classification via cost-based decision rules. In this context, it is crucial to understand that increasing prediction sensitivity always refers to both increasing \emph{and} decreasing confusion cost values. More precisely, the relative difference is critical according to \Cref{eq:cost-absolute}. Otherwise, confusions are valuated symmetrically as in the robotistic cost valuation, \cf \Cref{eq:simple-symm-cost}. Having now another look at the average cost matrices from the survey data in \Cref{fig:average-cost-matrices}, we realize that with respect to the class ``human'' the groups \emph{external} and \emph{female} have higher costs in the column as well as in the row compared to \emph{passenger} and \emph{male}, respectively. Whether such confusion costs valuation still translates to more human sensitive AI perception will be subject of the qualitative as well as consequential analysis in \Cref{sec:results-segmentation} and \Cref{sec:results-consequences}, respectively, after the statistical analysis of the survey data in the following.

\subsection{Statistical Evaluation of Different Confusion Cost Valuations} \label{sec: results-anova}

In order to study correlations between characteristics of the survey participants and their assessment of confusions, we statistically evaluate the survey data using an analysis of variance (ANOVA). We use this analysis technique to find differences in the means of data between two groups (one-way ANOVA).
As we deal with matrices (of size $6 \times 6$), which are not suitable for a standard $F$-test, we apply a modification thereof, see \Cref{app:f-test} for mathematical details. To put it another way, we perform a statistical test to analyze the degree of variability between two average cost matrices. 
To this end, the degree of variability is given by the $F$-statistic
\begin{equation}
    F := \frac{MS_B}{MS_W} = \frac{\textit{between-groups variance}}{\textit{within-groups variance}} \in \mathbb{R}_{\geq 0} ~.
\end{equation}
The greater the $F$-statistic, the greater the variance of the confusion valuations between different groups, implying that the examined groups differ more significantly.
After shuffling the data to randomly assign groups to any cost valuations, we determine how often the random $F$-statistic (denoted as $F_{random}$) is greater than the actual calculated $F$-statistic. In other words, we test the actual survey data against randomness. This approach is also known as bootstrapping \cite{moore14bootstrap} and provides the approximated $p$-value
\begin{equation}
    p := \frac{\# \{ F_\mathit{random} > F \}}{ \# \{ F_\mathit{random} \} } 
    = \frac{\textit{number of times } F_\mathit{random} > F}{\textit{total number of simulated } F_\mathit{random}} \in [0,1] ~,
\end{equation}
which indicates how likely two average cost matrices are drawn from the same distribution, \ie how likely there is \emph{no} difference between the examined groups with respect to their confusion cost valuations.

\begin{figure*}[t!]
    \centering
    \captionsetup[subfloat]{position=top}
    \subfloat[][\emph{passenger} \vs \emph{external}]{\includegraphics[width=0.45\textwidth]{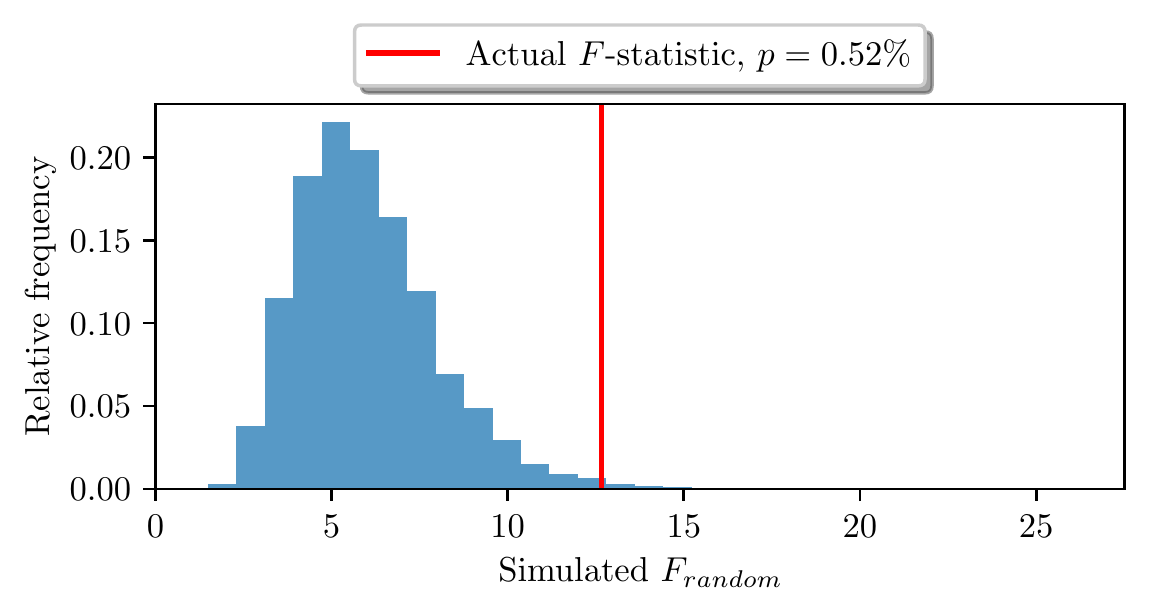}~~~~~~}
    \subfloat[][\emph{female} \vs \emph{male}]{\includegraphics[width=0.45\textwidth]{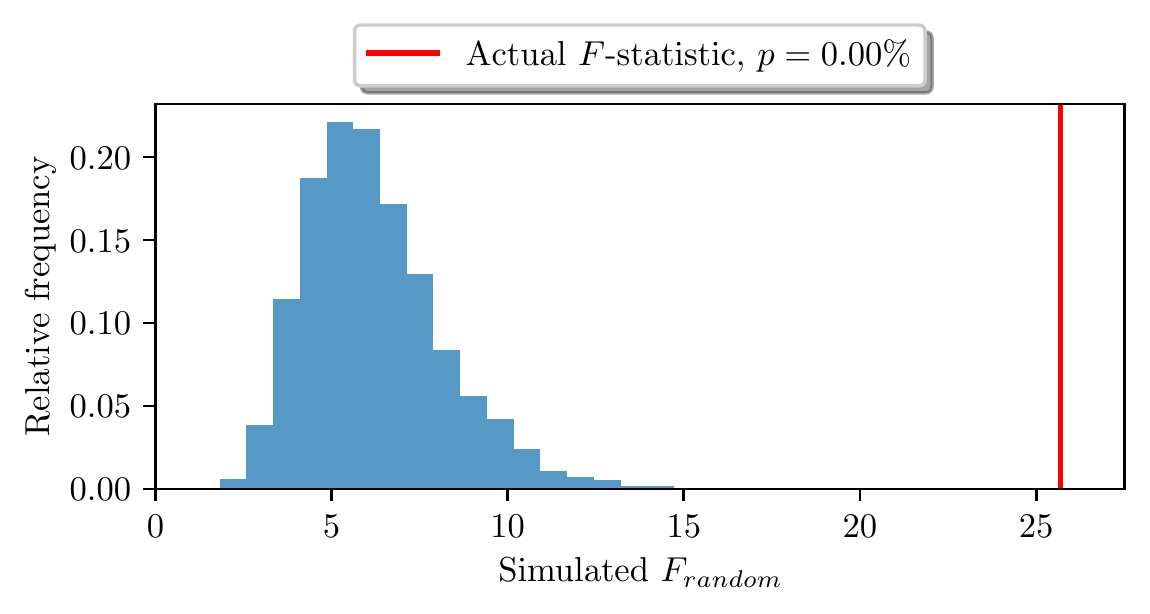}~~~~~~~}
    \caption{The distributions of simulated $F$-statistics, denoted by $F_\mathit{random}$, when the group affiliation of cost valuations in the survey data is randomly assigned. To this end, the survey dataset is shuffled 1M times in total. The greater the $F$-statistic, the greater the variance of the confusion cost valuations between the groups in the respective comparison. The red line corresponds to the actual calculated $F$-statistic according to the original, untouched survey data.}
    \label{fig:hist}
\end{figure*}

\begin{figure*}
    \centering
    \scalebox{0.9}{
    \begin{tikzpicture}

\def\w{-14}
\def\wp{-11.9}
\def\gapx{1.45}
\def\gapy{0.472}

\node [align=left,rotate=45,text width=1.65cm] at ({\wp+0*\gapx},8.2) {$\mathrm{``drivable"}$};
\node [align=left,rotate=45,text width=1.65cm] at ({\wp+1*\gapx},8.2) {$\mathrm{``nondrivable"}$};
\node [align=left,rotate=45,text width=1.65cm] at ({\wp+2*\gapx},8.2) {$\mathrm{``static"}$};
\node [align=left,rotate=45,text width=1.65cm] at ({\wp+3*\gapx},8.2) {$\mathrm{``info"}$};
\node [align=left,rotate=45,text width=1.65cm] at ({\wp+4*\gapx},8.2) {$\mathrm{``human"}$};
\node [align=left,rotate=45,text width=1.65cm] at ({\wp+5*\gapx},8.2) {$\mathrm{``dynamic"}$};

\node [align=left,text width=1.65cm, anchor=mid] at ({\wp+6*\gapx},7.2-0*\gapy) {$\mathrm{``drivable"}$};
\node [align=left,text width=1.65cm, anchor=mid] at ({\wp+6*\gapx},7.2-1*\gapy) {$\mathrm{``nondrivable"}$};
\node [align=left,text width=1.65cm, anchor=mid] at ({\wp+6*\gapx},7.2-2*\gapy) {$\mathrm{``static"}$};
\node [align=left,text width=1.65cm, anchor=mid] at ({\wp+6*\gapx},7.2-3*\gapy) {$\mathrm{``info"}$};
\node [align=left,text width=1.65cm, anchor=mid] at ({\wp+6*\gapx},7.2-4*\gapy) {$\mathrm{``human"}$};
\node [align=left,text width=1.65cm, anchor=mid] at ({\wp+6*\gapx},7.2-5*\gapy) {$\mathrm{``dynamic"}$};

\node [align=left,text width=1.65cm, anchor=mid] at ({\wp+6*\gapx},4.2-0*\gapy) {$\mathrm{``drivable"}$};
\node [align=left,text width=1.65cm, anchor=mid] at ({\wp+6*\gapx},4.2-1*\gapy) {$\mathrm{``nondrivable"}$};
\node [align=left,text width=1.65cm, anchor=mid] at ({\wp+6*\gapx},4.2-2*\gapy) {$\mathrm{``static"}$};
\node [align=left,text width=1.65cm, anchor=mid] at ({\wp+6*\gapx},4.2-3*\gapy) {$\mathrm{``info"}$};
\node [align=left,text width=1.65cm, anchor=mid] at ({\wp+6*\gapx},4.2-4*\gapy) {$\mathrm{``human"}$};
\node [align=left,text width=1.65cm, anchor=mid] at ({\wp+6*\gapx},4.2-5*\gapy) {$\mathrm{``dynamic"}$};

\node [align=left,text width=1.65cm, anchor=mid] at ({\wp+6*\gapx},1.2-0*\gapy) {$\mathrm{``drivable"}$};
\node [align=left,text width=1.65cm, anchor=mid] at ({\wp+6*\gapx},1.2-1*\gapy) {$\mathrm{``nondrivable"}$};
\node [align=left,text width=1.65cm, anchor=mid] at ({\wp+6*\gapx},1.2-2*\gapy) {$\mathrm{``static"}$};
\node [align=left,text width=1.65cm, anchor=mid] at ({\wp+6*\gapx},1.2-3*\gapy) {$\mathrm{``info"}$};
\node [align=left,text width=1.65cm, anchor=mid] at ({\wp+6*\gapx},1.2-4*\gapy) {$\mathrm{``human"}$};
\node [align=left,text width=1.65cm, anchor=mid] at ({\wp+6*\gapx},1.2-5*\gapy) {$\mathrm{``dynamic"}$};

\node [align=center] at (\w-2,3.15) {(b)};
\node [align=center] at (\w-.15,3) {$\Delta_{\substack{\mathit{female}\\\mathit{male}}}=$};
\node [align=center] at ({\w+5.2},3) 
{$\begin{pmatrix}
0      & 0.1131 & 0.3222 & 0.0829 & 0.0762 & 0.1193 \\
0.1733 & 0      & 0.1663 & 0.1228 & 0.0353 & 0.3988 \\
0.2960 & 0.0442 & 0      & 0.1706 & 0.0953 & 0.2013 \\
0.0282 & 0.1386 & 0.1776 & 0      & 0.0088 & 0.0028 \\
0.0127 & 0.0192 & 0.0787 & 0.0165 & 0      & 0.2112 \\
0.0574 & 0.1554 & 0.1380 & 0.0326 & 0.1427 & 0      
\end{pmatrix}$};
\node [align=center] at (\w+14.6,3) {$\Sigma_{\substack{\mathit{female}\\\mathit{male}}}=3.6394$};

\node [align=center] at (\w-2,6.15) {(a)};
\node [align=center] at (\w-.3,6) {$\Delta_{\substack{\mathit{passenger}\\\mathit{external}}}=$};
\node [align=center] at ({\w+5.2},6) 
{$\begin{pmatrix}
0      & 0.1313 & 0.1377 & 0.1971 & 0.2054 & 0.3032 \\
0.1054 & 0      & 0.3394 & 0.1407 & 0.0097 & 0.1528 \\
0.0655 & 0.1746 & 0      & 0.0267 & 0.0275 & 0.0761 \\
0.1303 & 0.1245 & 0.0939 & 0      & 0.0768 & 0.0046 \\
0.0098 & 0.3672 & 0.2676 & 0.0319 & 0      & 0.0895 \\
0.0207 & 0.3129 & 0.1524 & 0.1831 & 0.0622 & 0    
\end{pmatrix}$};
\node [align=center] at (\w+14.5,6) {$\Sigma_{\substack{\mathit{passenger}\\\mathit{external}}}=4.0221$};

\node [align=center] at (\w-2,.15) {(c)};
\node [align=center] at (\w-.3,0) {$\Delta_{\substack{\mathit{all\,survey}\\\mathit{robot}}}=$};
\node [align=center] at ({\w+5.2},0) 
{$\begin{pmatrix}
0      & 0.4195 & 0.0366 & 0.1705 & 0.6560 & 0.2055 \\
1.2728 & 0      & 1.1589 & 0.5589 & 0.3640 & 0.3462 \\
0.7875 & 0.7304 & 0      & 0.5362 & 0.4256 & 0.3986 \\
0.5388 & 0.3456 & 0.8982 & 0      & 0.0495 & 0.0855 \\
1.6849 & 0.1799 & 0.1685 & 0.2839 & 0      & 0.2705 \\
0.9179 & 0.1568 & 0.2481 & 0.2635 & 0.2434 & 0   
\end{pmatrix}$};

\node [align=center] at (\w+14.55,0) {$\Sigma_{\substack{\mathit{all\,survey}\\\mathit{robot}}}=\mathbf{14.4036}$};

\end{tikzpicture}
    }
    \caption{Value-wise differences of confusion costs assessment between different groups. Here, $\Delta$ denotes the difference matrix of confusion cost values for the comparisons of the groups (a) \emph{passenger of the self-driving car} \vs \emph{external traffic participant}, (b) \emph{female} \vs \emph{male}, and (c) \emph{all survey participants} \vs \emph{robotistic cost valuation / Bayes decision rule}. Furthermore, $\Sigma$ denotes the sum of entries in the difference matrices of the respective comparisons.}
    \label{fig:difference-cost-matrices}
\end{figure*}

In our analysis, we test for differences when comparing the groups \emph{passenger of the self-driving car} \vs \emph{external traffic participant} and \emph{female \vs male}. The results of our conducted statistical tests are summarized in \Cref{fig:hist}.
Regarding the evaluation of the perspective the $p$-value is $p=0.52\%$, while regarding the gender $p=0.00\%$. Such small $p$-values indicate that there is significant evidence for statistical differences between the average cost matrices associated with the examined groups. This is particularly significant when comparing the groups \emph{female} \vs \emph{male} since in our analysis there has not been a single randomly simulated $F$-statistic that exceeds the actual calculated $F$-statistic. Based on the findings of these statistical tests, we conclude that the characteristics related to the assigned perspective as well as the gender of the survey participants both likely have an impact on their respective confusion costs valuation. 

However, the evaluation methodology in this subsection does not provide information regarding by what amount and for which specific type of confusion the costs differ. To this end, we further investigate the differences between average cost valuations for specific types of confusion. In particular, we additionally compare the average confusion cost matrix of \emph{all survey participants} against the \emph{robotistic confusion costs valuation}, \ie constant costs for each possible type of confusion, \cf \Cref{eq:simple-symm-cost}. The difference matrices are given in \Cref{fig:difference-cost-matrices}. The previously conducted statistical tests might suggest that the difference in the confusion cost valuation between different groups of survey participants tend to be significant, but we observe that the difference between all survey participants as unit and the robotistic cost valuation is even more drastic. The total sum of entries in the corresponding difference matrix yields 14.40 compared to 4.02 and 3.69 for the perspective and gender, respectively. This result let us conclude that the subjective sense of the survey participants with respect to confusion costs in the context of street scenes \emph{clearly disagrees with the robotistic confusion costs valuation}, and thus with the Bayes decision rule, which is used by default in any machine learning model.

\subsection{Qualitative Evaluation of Different Confusion Cost Matrices} \label{sec:results-segmentation}

\begin{table}[]
    \centering
    \begin{tabular}{c||ccc|ccc}
        survey group \textbackslash\ performance metric & mean IoU & mean recall & mean precision & human IoU & human recall & human precision \\
        \hline\hline
        {passenger of the self-driving car} & 82.3          & 94.7          & 85.9          & 79.9          & \textbf{94.6} & 85.0          \\
        {external traffic participant}      & 81.1          & 93.9          & 85.1          & 82.8          & 93.7          & 86.2          \\
        \hline                                                      
        {female survey participants}        & 80.9          & 94.1          & 84.8          & 82.9          & 94.0          & 86.0          \\
        {male survey participants}          & 82.0          & 94.6          & 85.8          & 82.7          & 94.2          & 85.6          \\
        \hline                                                      
        {all survey participants}           & 81.6          & 94.4          & 85.3          & 82.7          & 94.3          & 85.5          \\
        {robot / Bayes decision rule}       & \textbf{90.0} & \textbf{95.2} & \textbf{94.0} & \textbf{84.1} & 91.8          & \textbf{88.2}
    \end{tabular}
    \caption{Semantic segmentation performance on the Cityscapes validation data when incorporating different costs of confusion provided by different groups. The most commonly used performance metric for this computer vision task is the \emph{mean IoU}, which is the class-wise intersection over union (IoU) averaged over all available classes. For further insights, particularly with respect to performance tradeoffs (\cf \Cref{sec:Tradeoffs}) we additionally include the metrics \emph{recall} as well \emph{precision} (\cf \Cref{fig:rec-prc}), and also restrict the evaluation to the human class only. Note that the values in this table denote percentages (\%). For all performance metrics higher values indicate better semantic segmentation performance with 100\% being the score of a perfect classifier.}
    \label{tab:seg-performance}
\end{table}

\begin{figure*}
    \centering
    \subfloat[passenger of the self-driving car]{\includegraphics[width=0.49\textwidth]{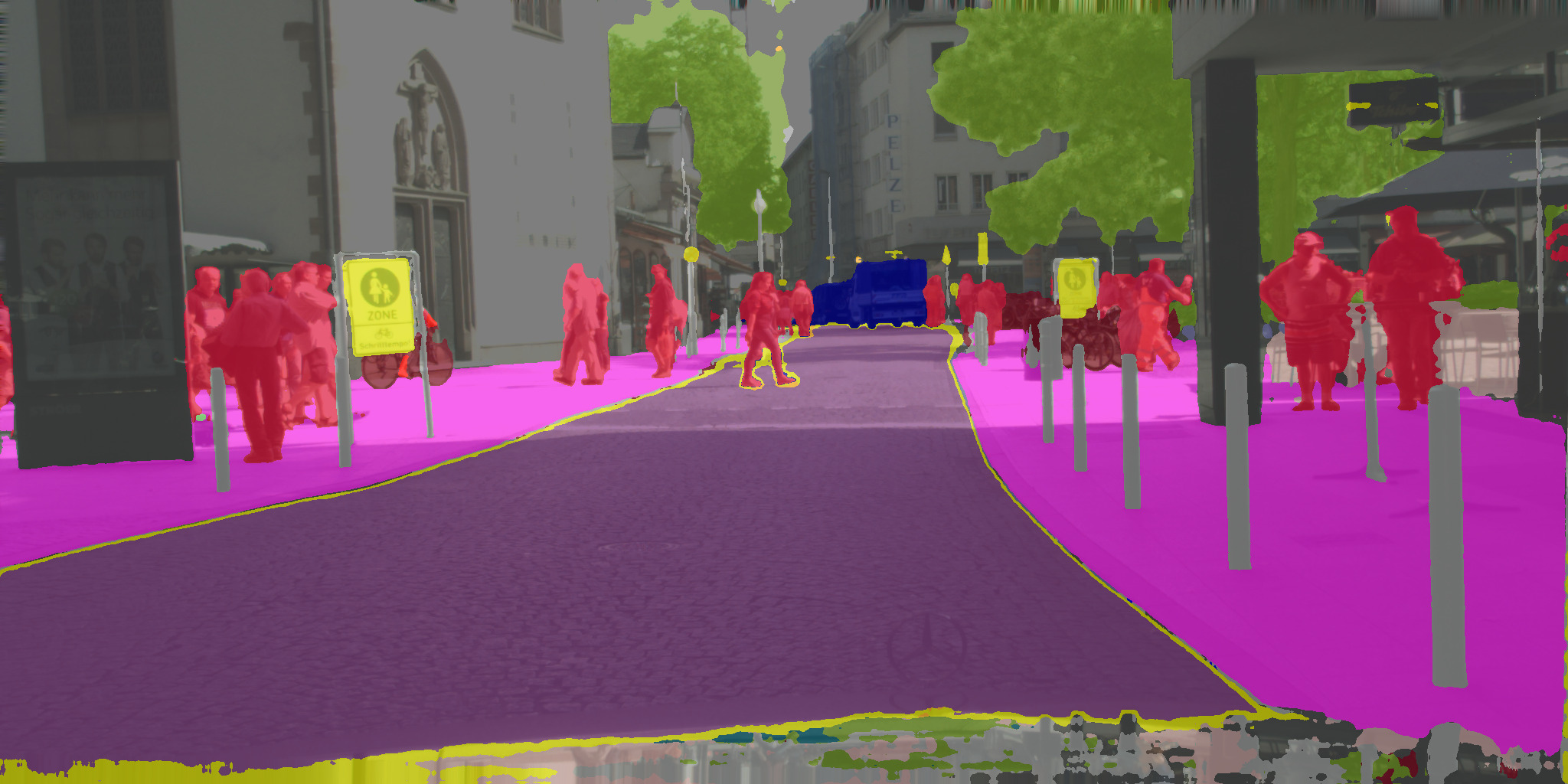}}~
    \subfloat[external traffic participant]{\includegraphics[width=0.49\textwidth]{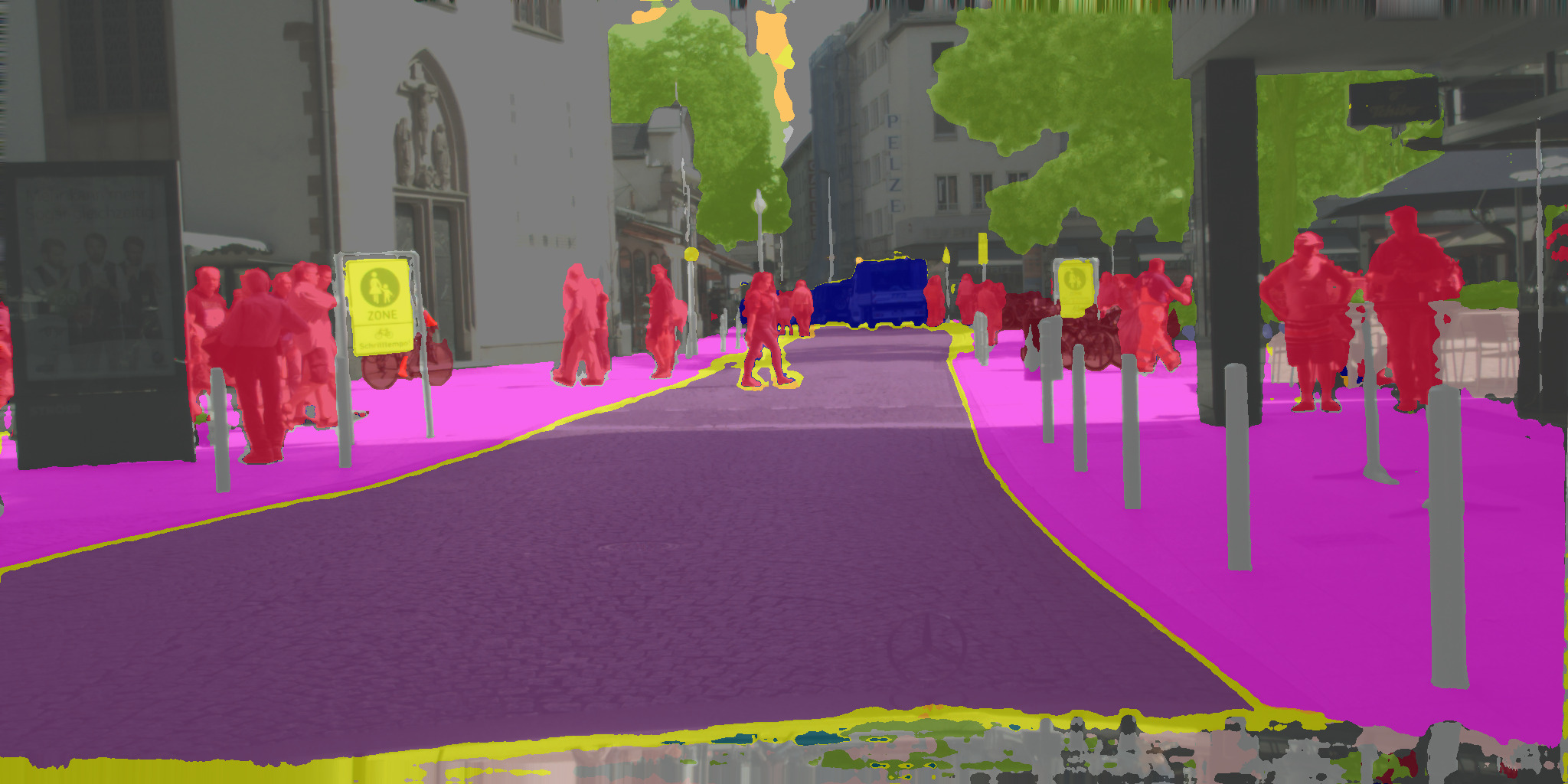}} \\
    \subfloat[female survey participant]{\includegraphics[width=0.49\textwidth]{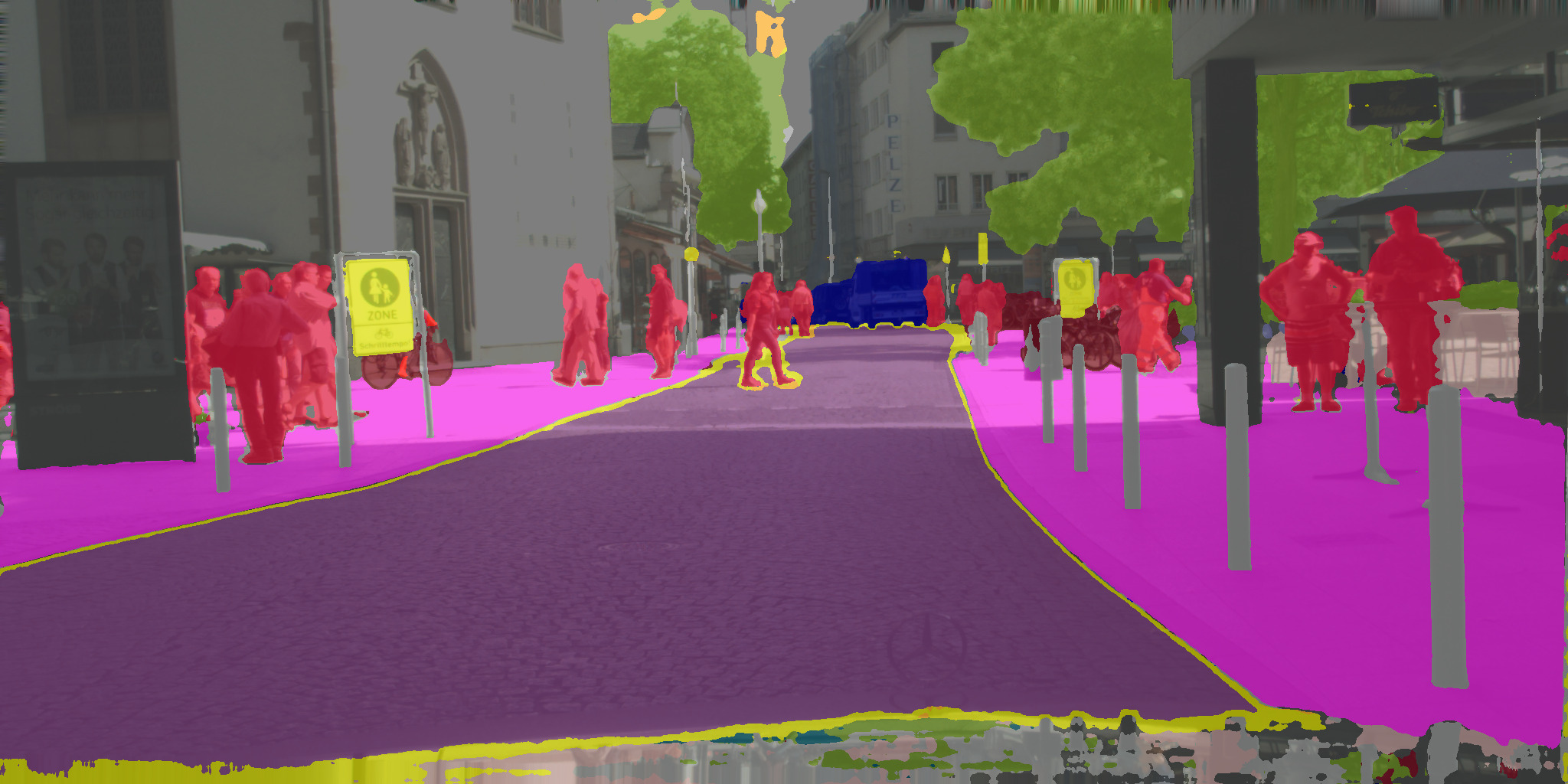}}~
    \subfloat[male survey participant]{\includegraphics[width=0.49\textwidth]{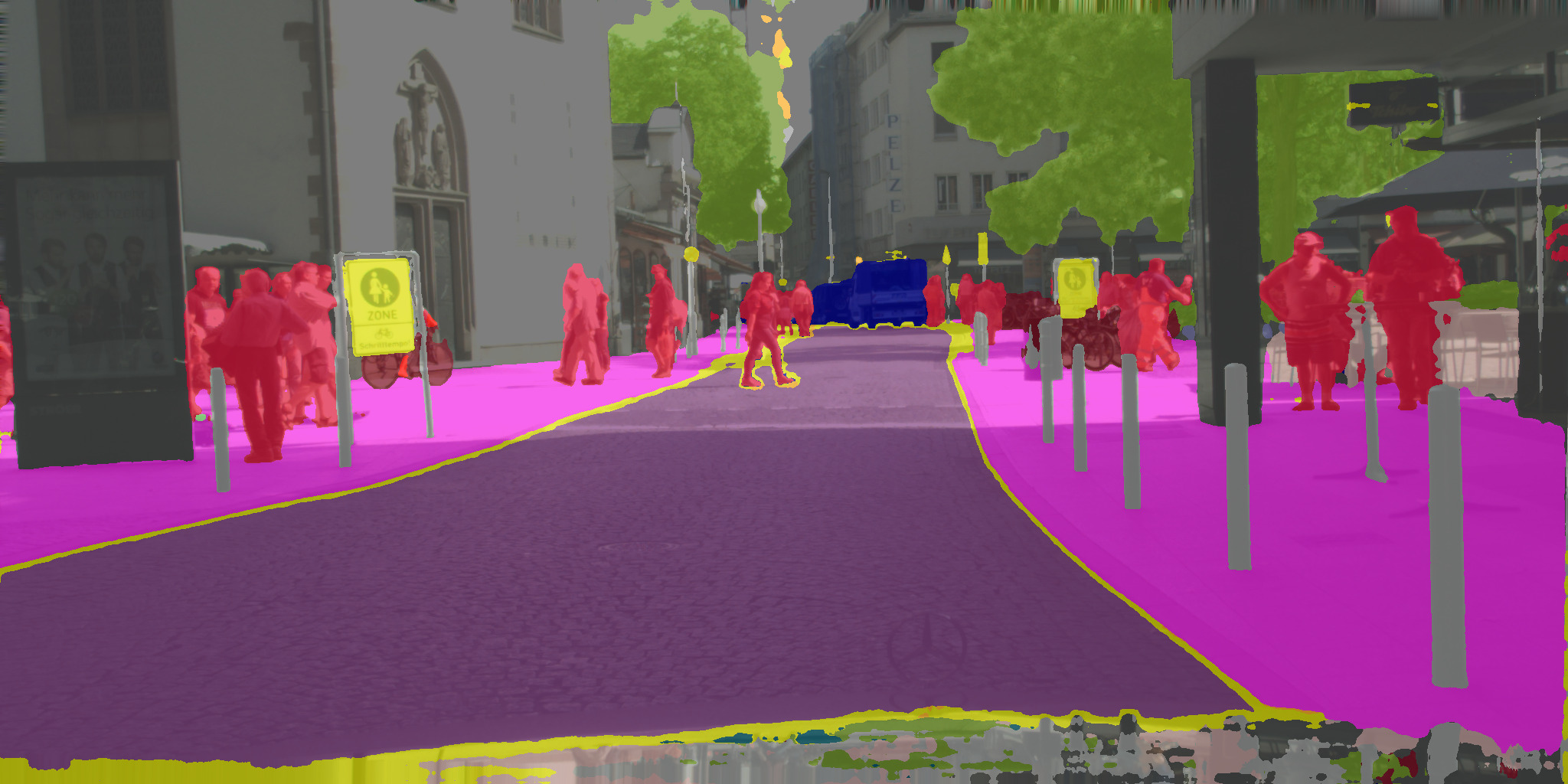}} \\
    \subfloat[all survey participants]{\includegraphics[width=0.49\textwidth]{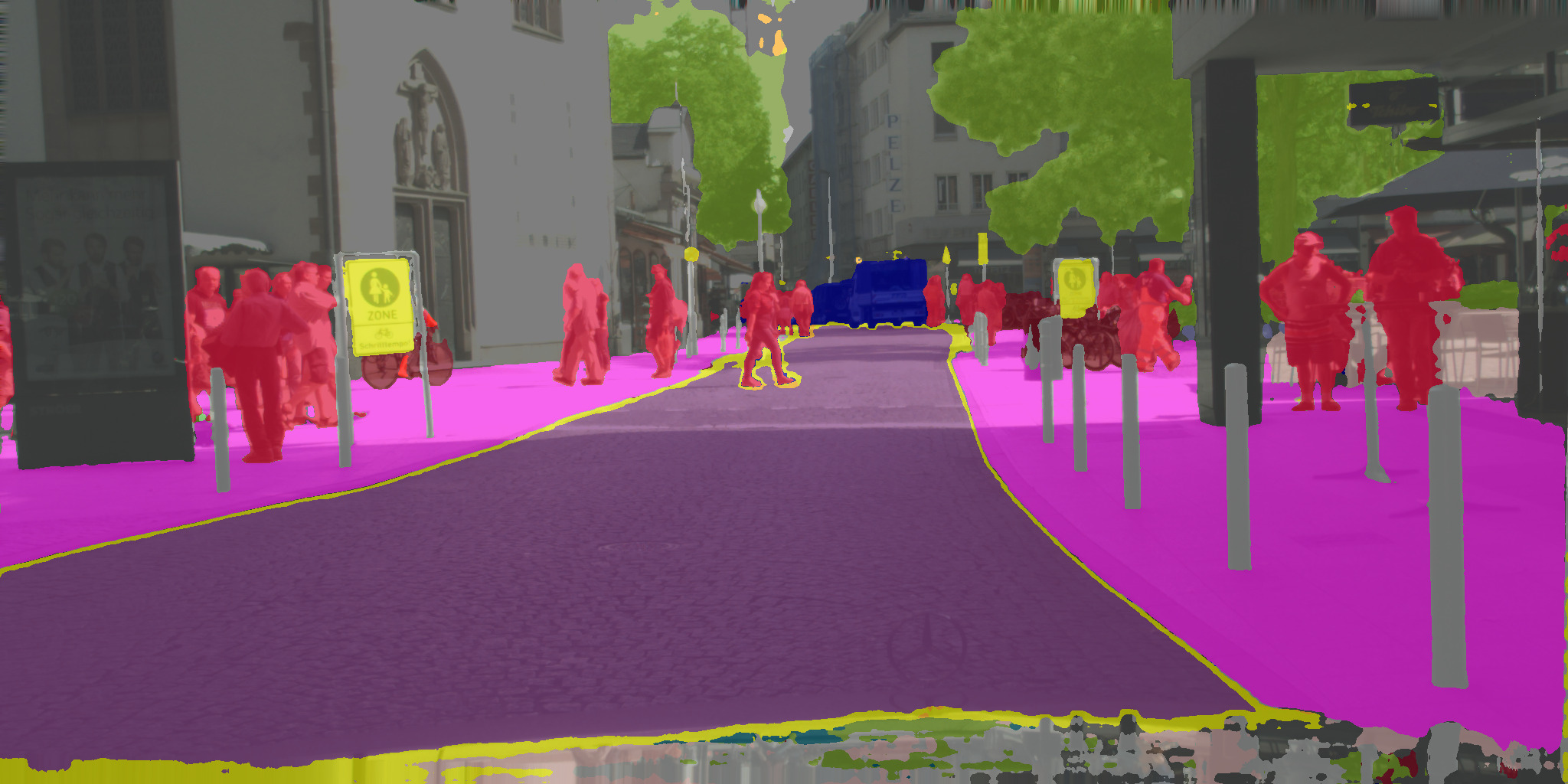}}~
    \subfloat[robot / Bayes decision rule]{\includegraphics[width=0.49\textwidth]{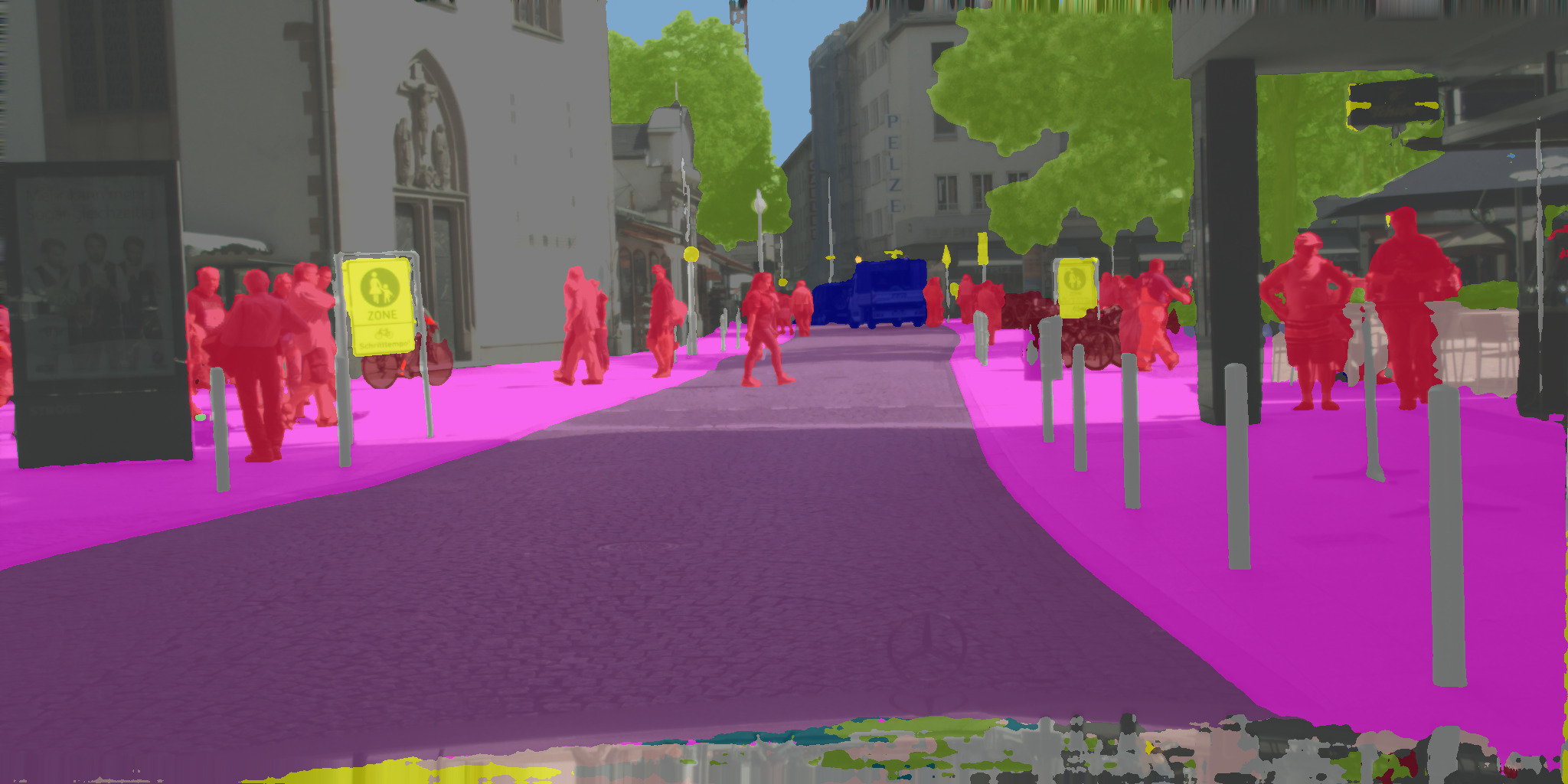}} \\
    \caption{Changes in AI perception of an urban street scene by using different confusion cost matrices, \cf also \Cref{fig:average-cost-matrices}, in the cost-based decision rule integrated within a state-of-the-art deep neural network for semantic segmentation.}
    \label{fig:seg-comparison}
\end{figure*}
In this subsection, we investigate the actual changes in AI perception of street scenes obtained by using the confusion cost matrices from the survey, \cf \Cref{fig:average-cost-matrices}.
We test the quality of AI perception by integrating the given confusion cost matrices into the cost-based decision rule within a deep neural network (DNN) for semantic segmentation, \cf \Cref{eq:exp-cost-sum}. In this context, semantic segmentation can be described as pixel-wise image classification task, \ie assigning an object category to each single pixel of an input image, and thus providing the finest level of detection and localization of objects in scenes. As underlying semantic segmentation DNN we employ the state-of-the-art \texttt{DeepLabV3+} model with \texttt{WideResNet38} backbone trained by Nvidia \cite{Zhu2019}.

We report numerical semantic segmentation performance results in \Cref{tab:seg-performance}. As expected, using the Bayes decision rule yields the best overall segmentation performance. The commonly used performance metric mean IoU drops by 8.4 percent points when using the confusion costs from the survey. Among the examined survey groups, the group \emph{external} achieves the best score with 82.3\%. Restricting the evaluation on human classification only, the Bayes decision rule still performs best, but by smaller margin of 1.4 percent points over the cost valuations of all survey participants. Among the survey groups, the group \emph{female} achieves the best score with 82.9\%. Noteworthy, all groups improve recall over the Bayes decision rule by up to 2.8 percent points for the group \emph{passenger}.

For visual inspection, we provide exemplary semantic segmentation masks for one scene of the Cityscapes dataset \cite{cityscapes} in \Cref{fig:seg-comparison}. In general, we realize that all semantic segmentation masks look similar in large parts. In particular with respect to human classification, differences in the outputs are barely visible. The small changes compared to the standard Bayes decision rule include slightly larger segment predictions for the class human, which is a consequence of the increased sensitivity towards human predictions already discussed in \Cref{sec: results-matrices}. Between the examined survey groups the visual differences in the semantic segmentation masks are marginal, particularly in regard to safety relevant classes such as ``road'', ``info'' and ``human'', which let us conclude that the cost matrices provided by the survey participants do not significantly impact the segmentation quality via deep neural networks.

\subsection{Consequential Evaluation of Different Confusion Cost Matrices} \label{sec:results-consequences}

\begin{figure*}
    \centering
    \begin{tikzpicture}
        \draw (0,15) rectangle (5,0); 
        \draw (0,15) rectangle (10,5);
        \draw (0,15) rectangle (15,10);
        \node (external_passenger) at (2.4,12.8) {\includegraphics[scale=0.42]{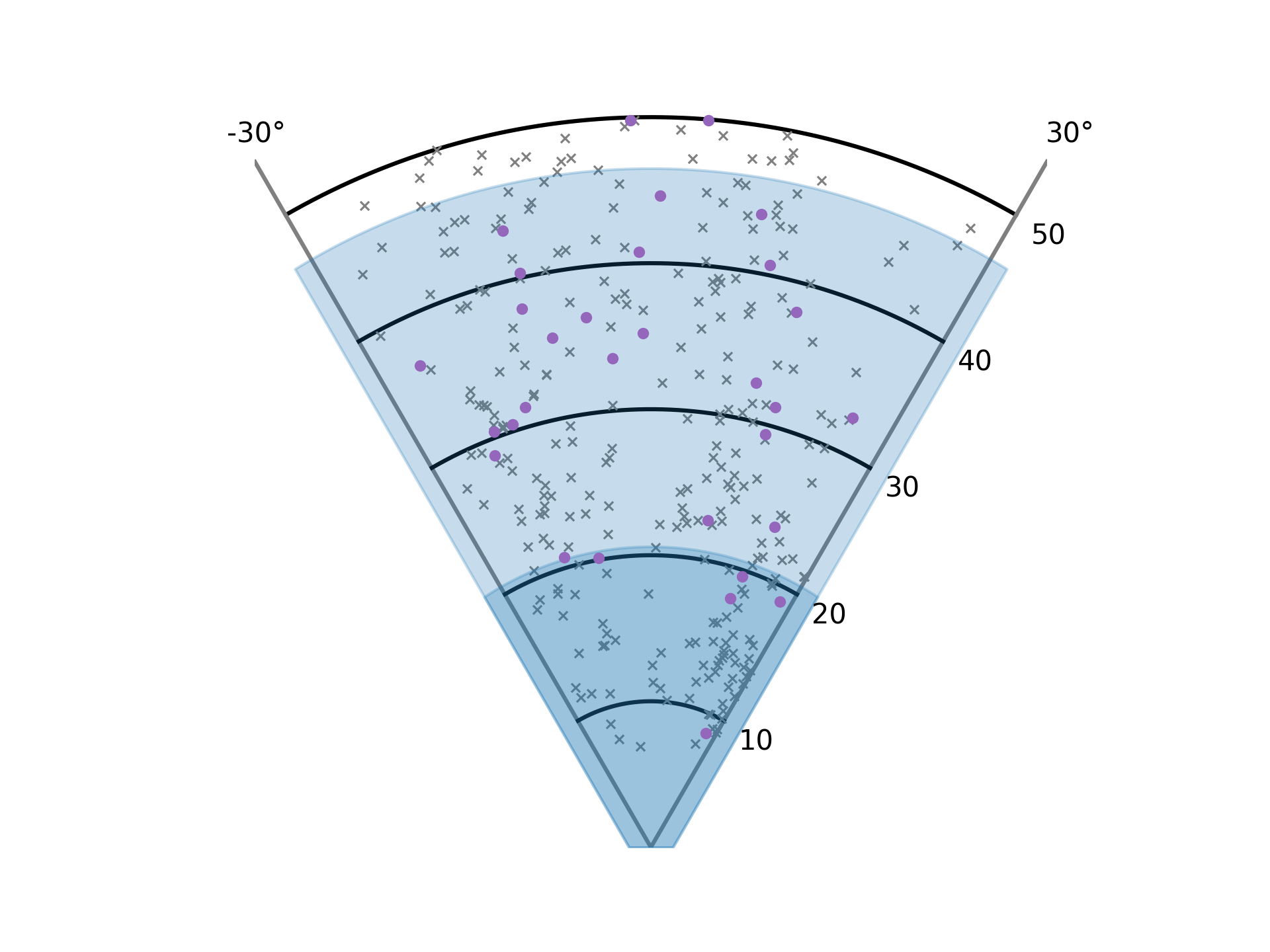}};
        \node (external_robot) at (7.4,12.8) {\includegraphics[scale=0.42]{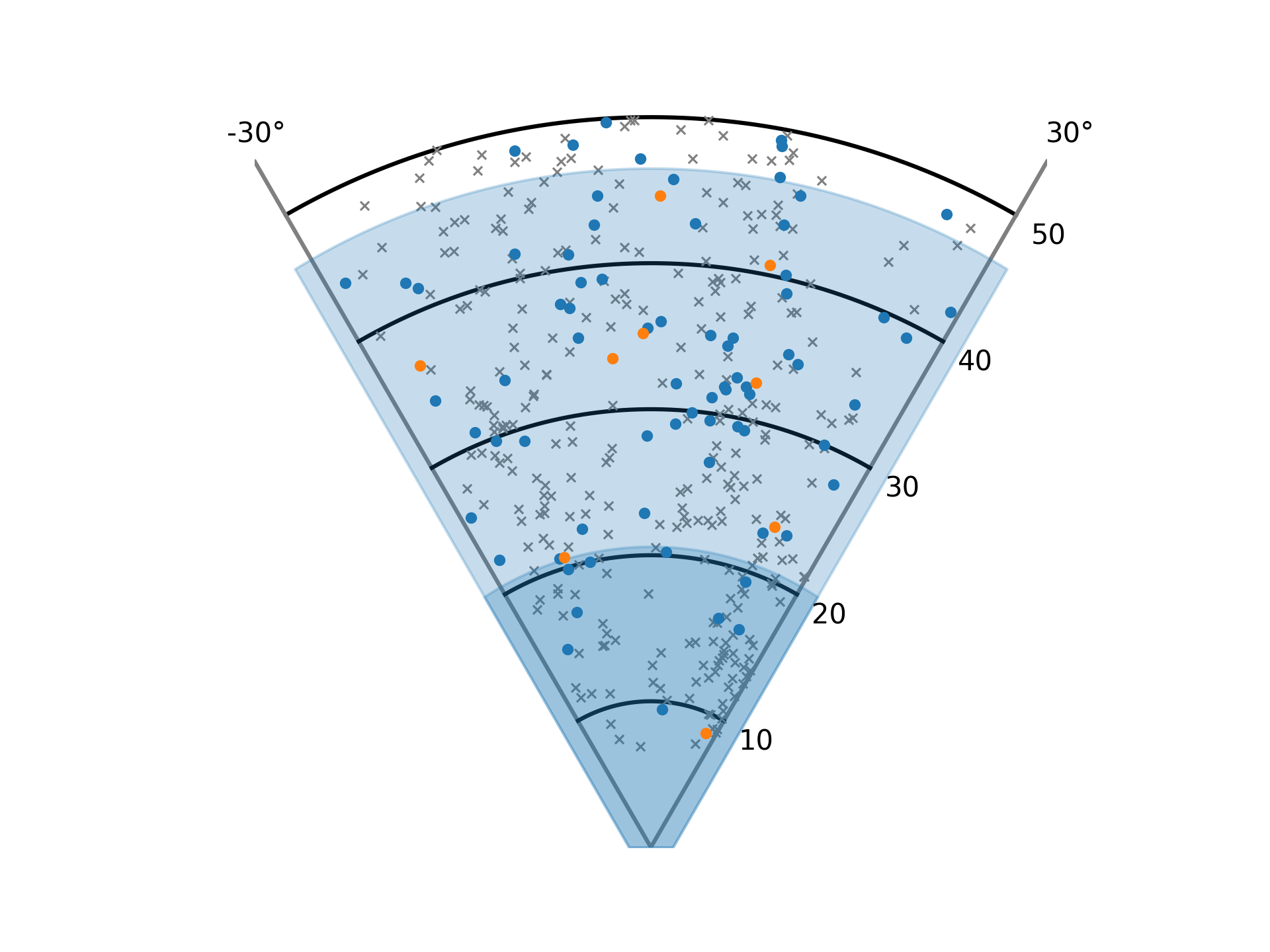}};
        \node (passenger_robot) at (2.4,7.8) {\includegraphics[scale=0.42]{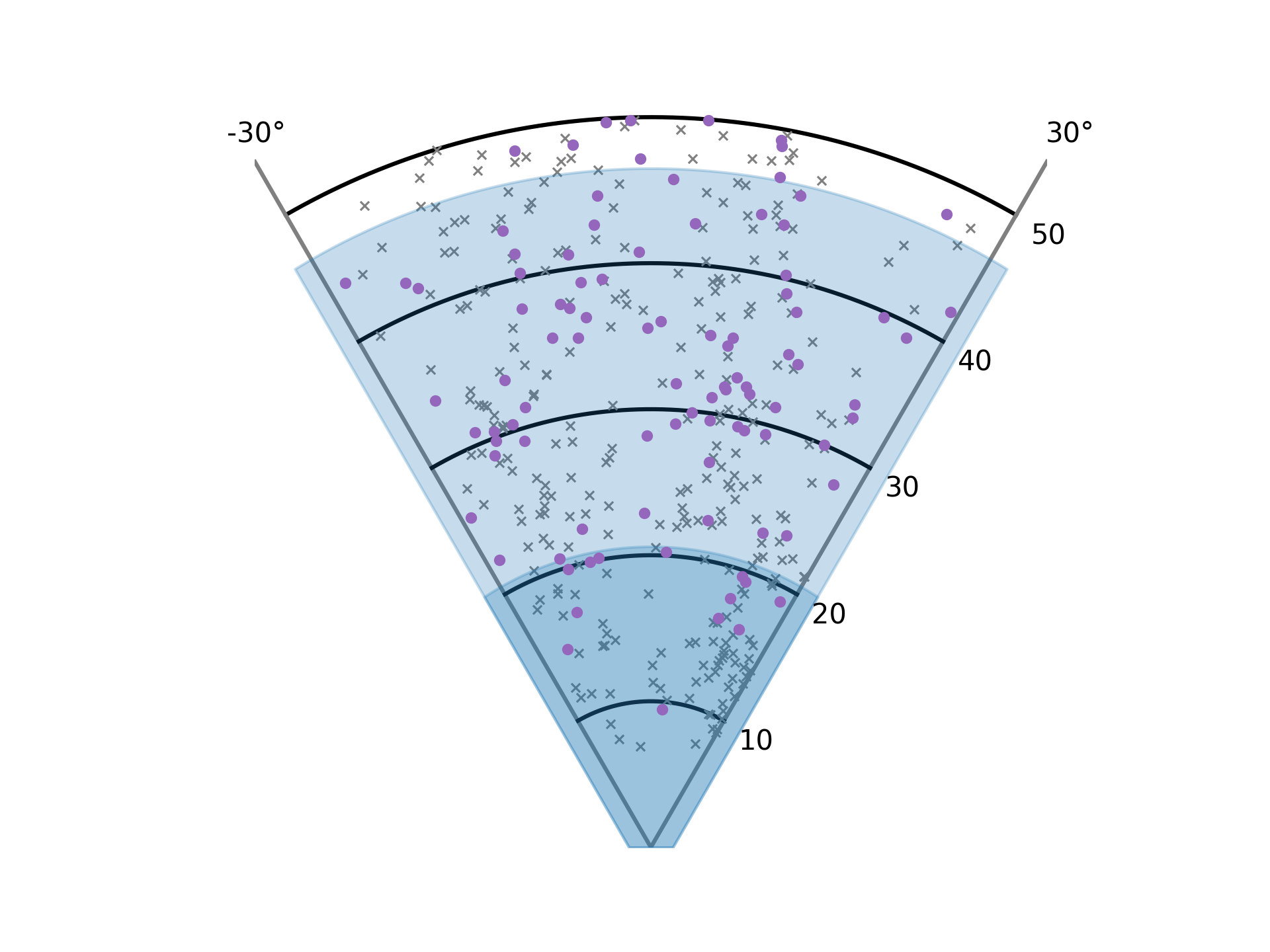}};
        \node (passenger_pie) at (2.4,2.7) {\includegraphics[scale=0.42]{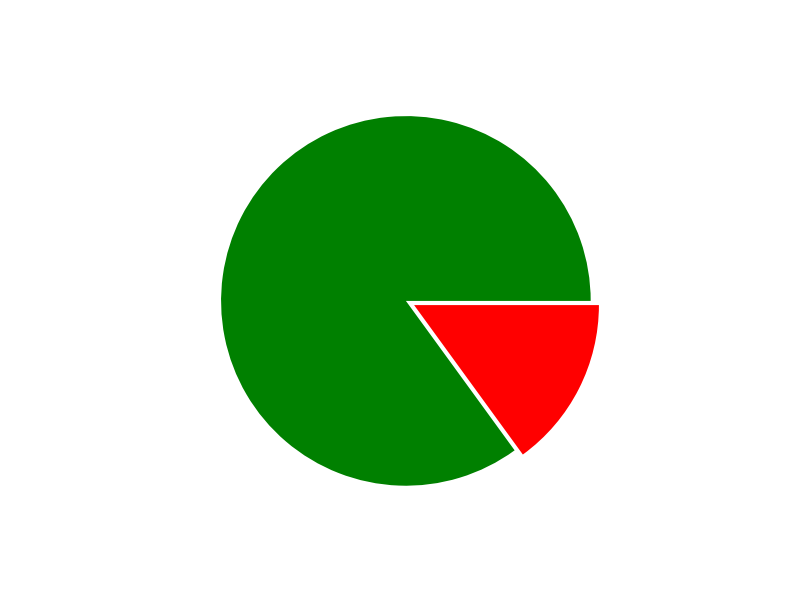}};
        \node (external_pie) at (12.4,12.7) {\includegraphics[scale=0.42]{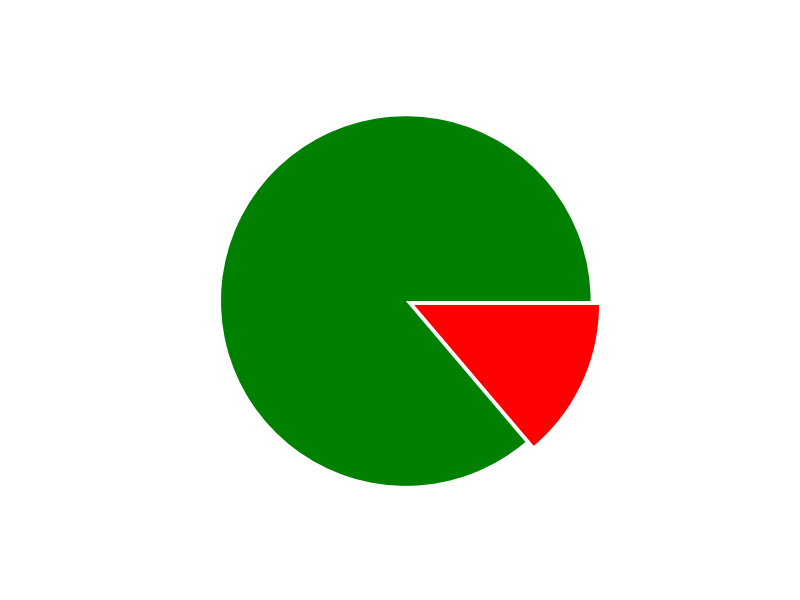}};
        \node (robot_pie) at (7.4,7.7) {\includegraphics[scale=0.42]{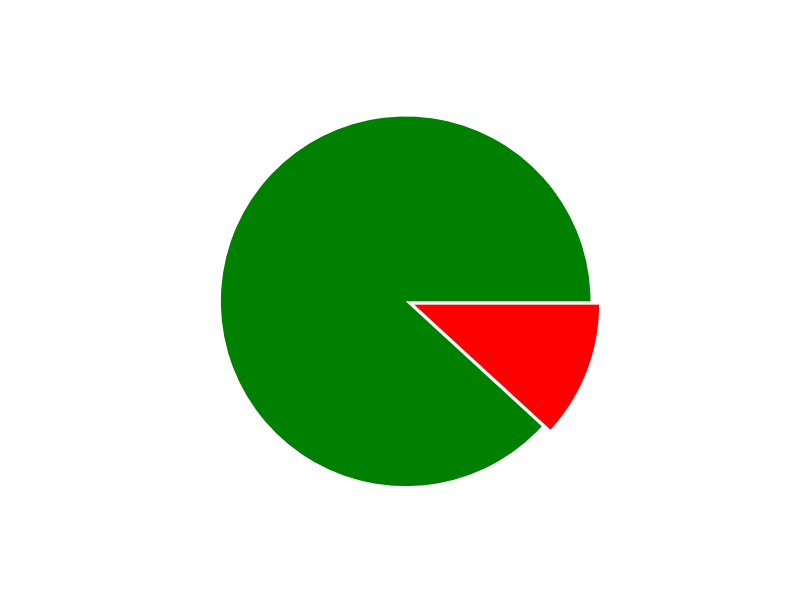}};
        \node (passenger_prc) at (2.5,0.5) [align=center] {\small $ \color{Green} \mathrm{TP} = 85.01 \%  ~~~~~ \color{Red} \mathrm{FP} = 14.99 \%$\\ total: 15,892,502 pixels};
        \node (robot_prc) at (7.5,5.5) [align=center] {\small $ \color{Green} \mathrm{TP} = 88.18 \%  ~~~~~ \color{Red} \mathrm{FP} = 11.82 \%$\\ total: 14,873,059 pixels};
        \node (external_prc) at (12.5,10.5) [align=center] {\small $ \color{Green} \mathrm{TP} = 86.20 \%  ~~~~~ \color{Red} \mathrm{FP} = 13.80 \%$\\ total: 15,506,832 pixels};
        \node (label_robot) at (-0.4,7.4) [rotate=90, align=center] {\includegraphics[height=8px]{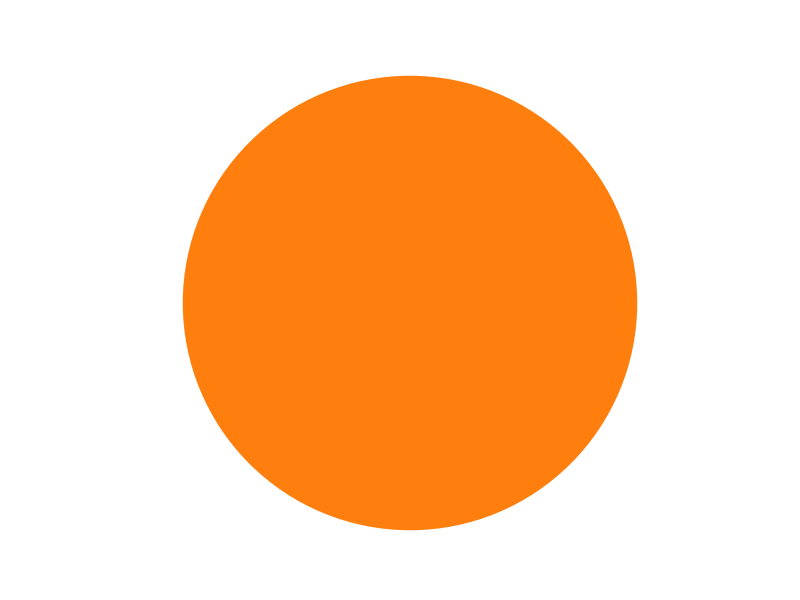} robot};
        \node (label_external) at (-0.4,12.4) [rotate=90, align=center] {\includegraphics[height=8px]{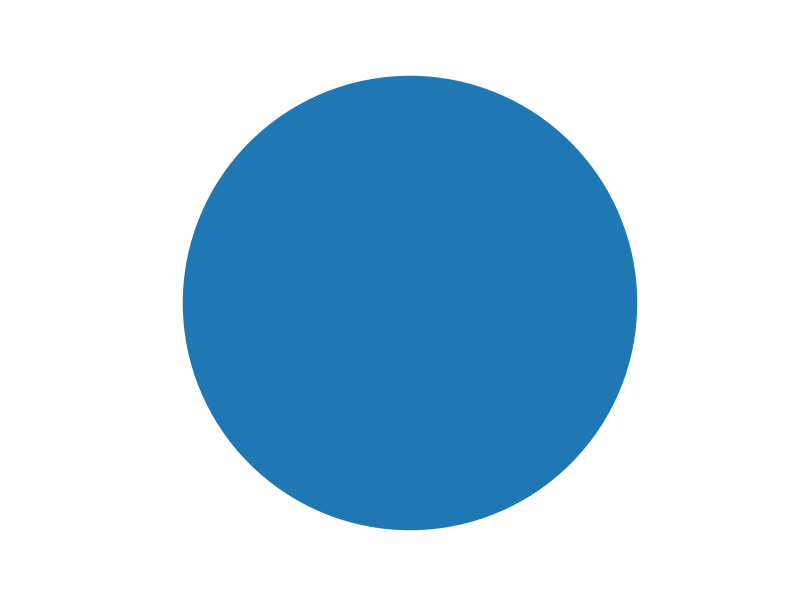} external};
        \node (label_passenger) at (-0.4,2.4) [rotate=90, align=center] {\includegraphics[height=8px]{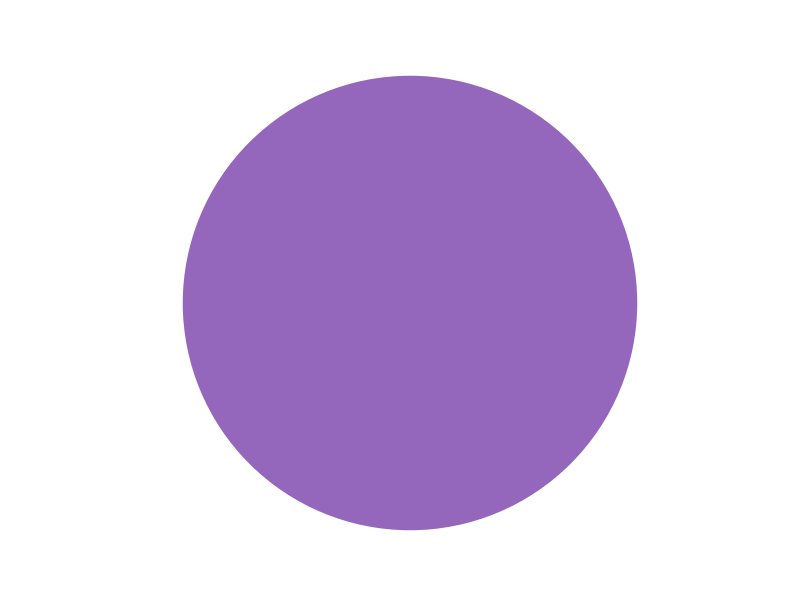} passenger};
        \node (label_robot) at (7.4,15.4) [align=center] {\includegraphics[height=8px]{C1.png} robot};
        \node (label_external) at (12.4,15.4) [align=center] {\includegraphics[height=8px]{C0.png} external};
        \node (label_passenger) at (2.4,15.4) [align=center] {\includegraphics[height=8px]{C4.png} passenger};
        \node [car top,draw=black!,fill=black!90,minimum width=.8cm,rotate=90] at (2.485,10.5) {};
        \node [car top,draw=black!,fill=black!90,minimum width=.8cm,rotate=90] at (7.485,10.5) {}; 
        \node [car top,draw=black!,fill=black!90,minimum width=.8cm,rotate=90] at (2.485,5.5) {};
    \end{tikzpicture}
    \caption{Consequential comparisons between the cost assignments of the groups \emph{passenger}, \emph{external}, and \emph{robot} with respect to the detection of humans. On the diagonal, the pie charts indicate the precision of human predictions as practicability measure for the respective cost assignments. The graphs on the off diagonal display the perceivable area ahead of the ego-vehicle from a bird’s eye perspective. Here, gray crosses show human instances that are overlooked by both considered groups in the comparison, whereas colored dots show human instances that are detected only by the group corresponding to the respective color.}
    \label{fig:passenger-external-robot}
\end{figure*}

\begin{figure*}
    \centering
    \begin{tikzpicture}
        \draw (0,15) rectangle (5,0); 
        \draw (0,15) rectangle (10,5);
        \draw (0,15) rectangle (15,10);
        \node (male_female) at (2.4,12.8) {\includegraphics[scale=0.42]{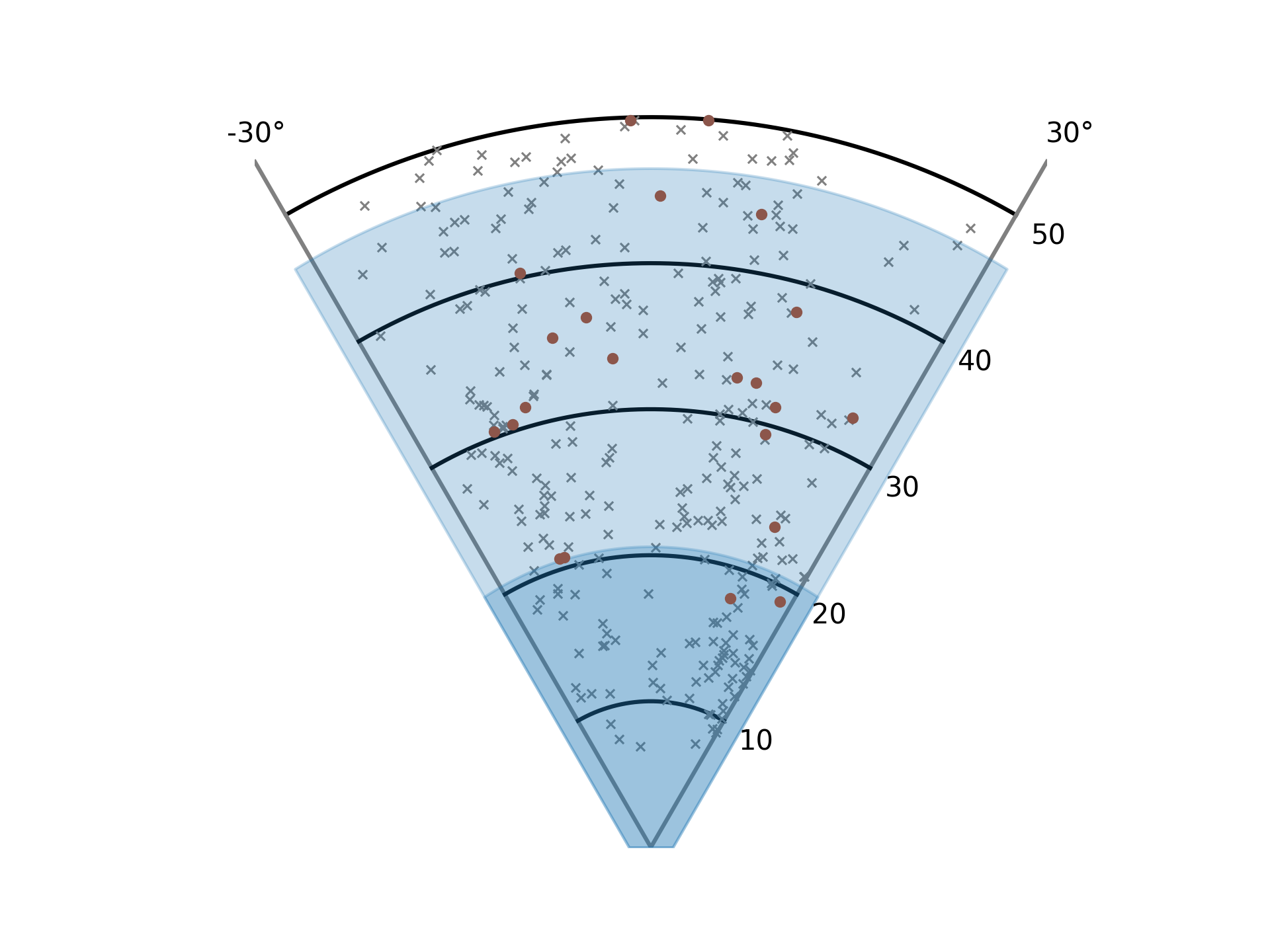}};
        \node (female_robot) at (2.4,7.8)  {\includegraphics[scale=0.42]{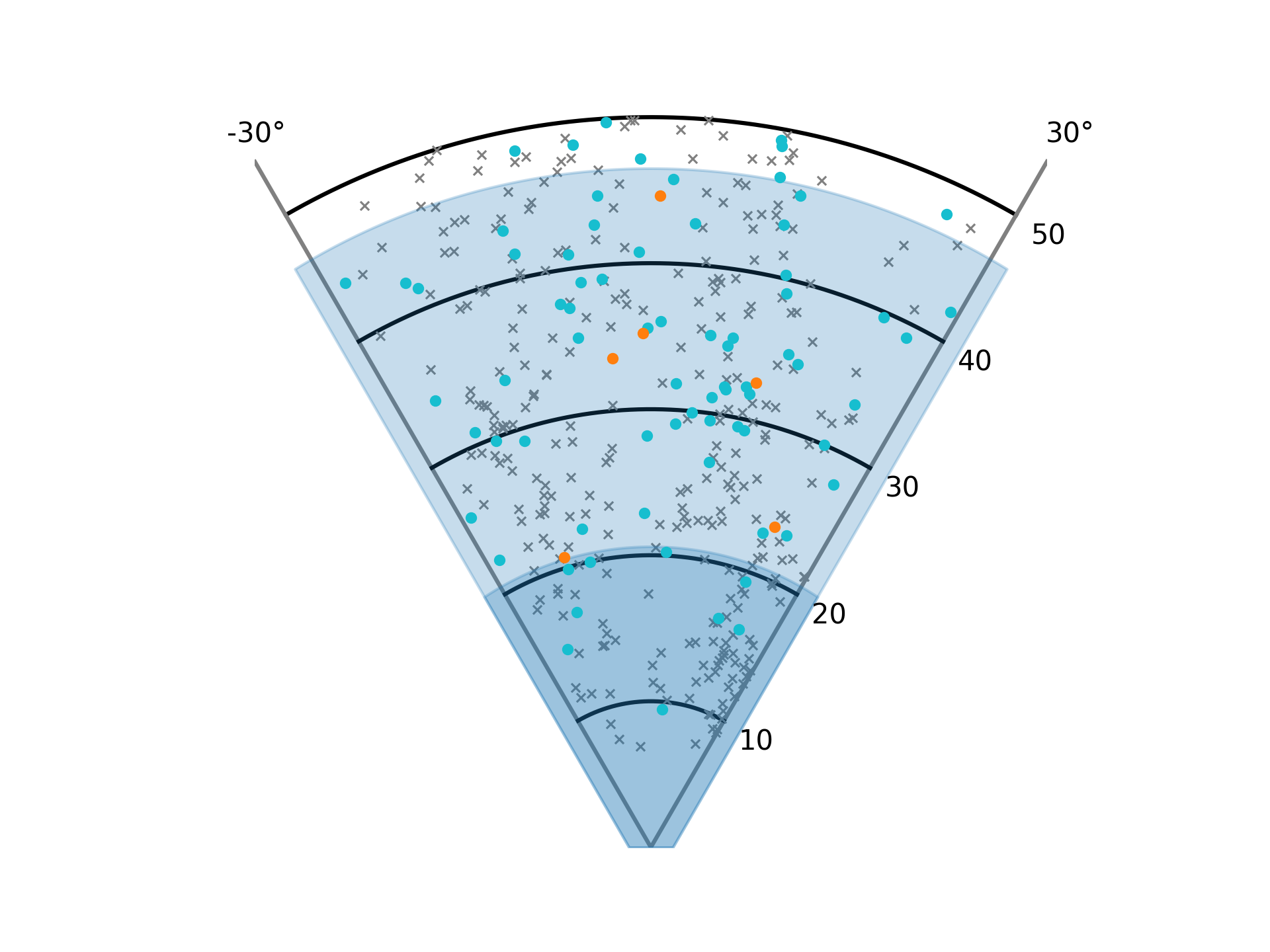}}; 
        \node (male_robot) at (7.4,12.8) {\includegraphics[scale=0.42]{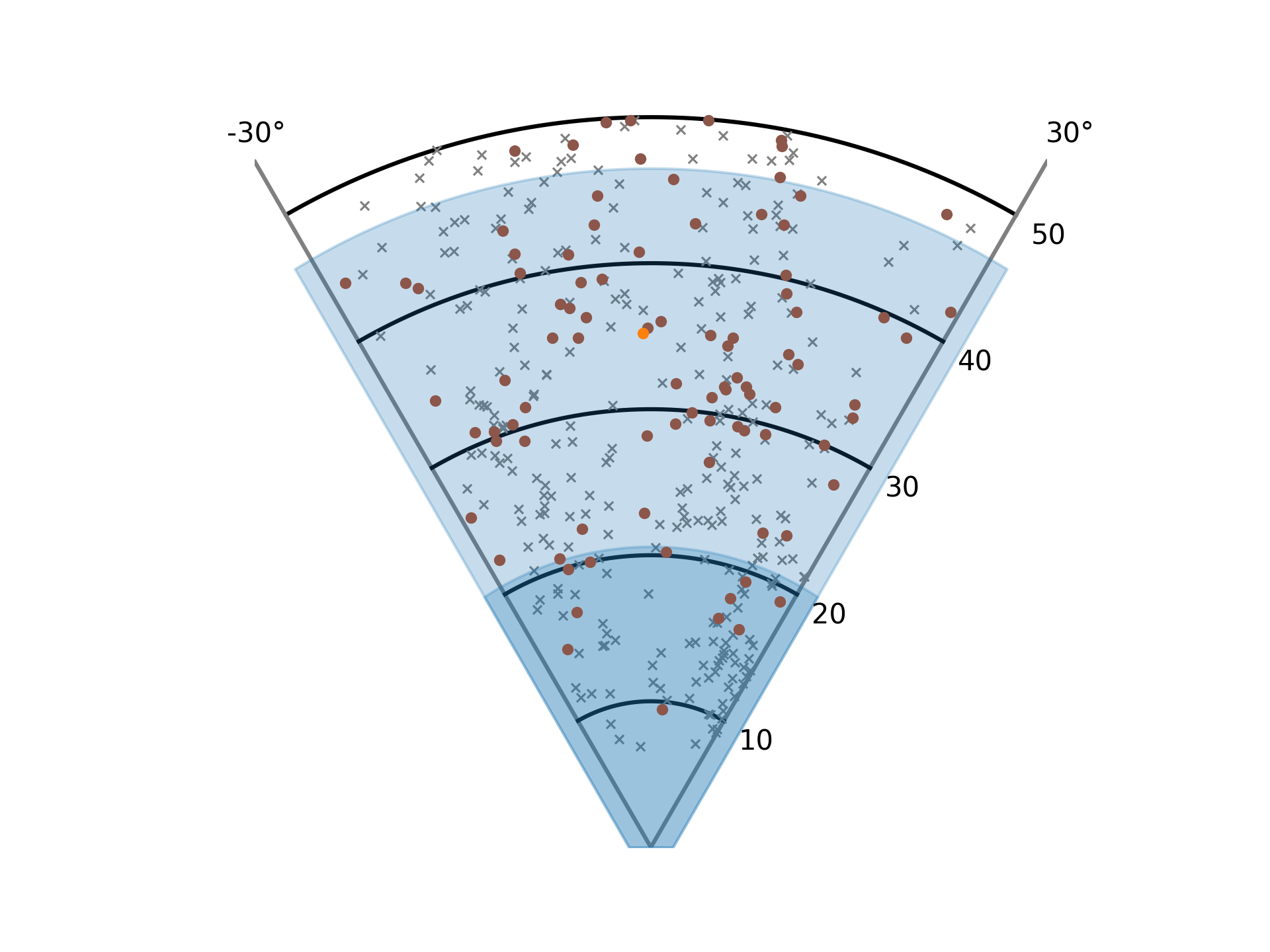}};
        \node (female_pie) at (2.4,2.7) {\includegraphics[scale=0.42]{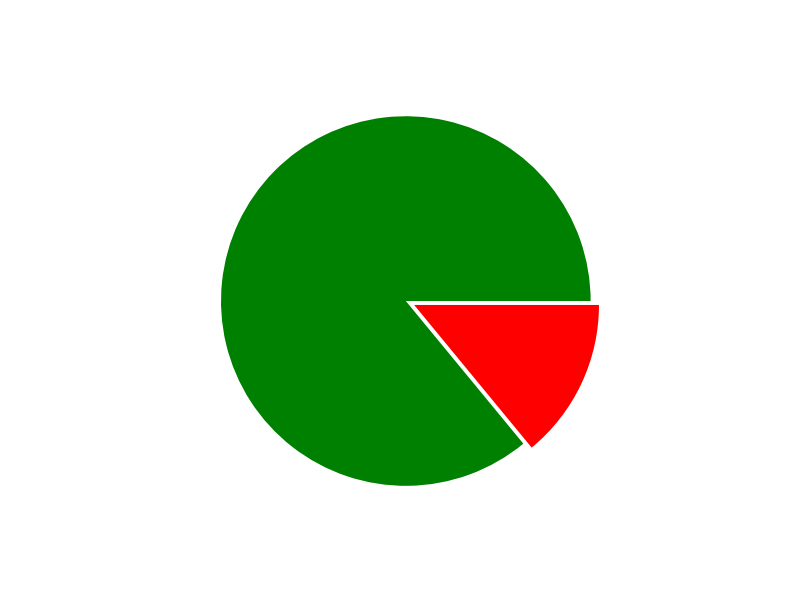}};
        \node (male_pie) at (12.4,12.7) {\includegraphics[scale=0.42]{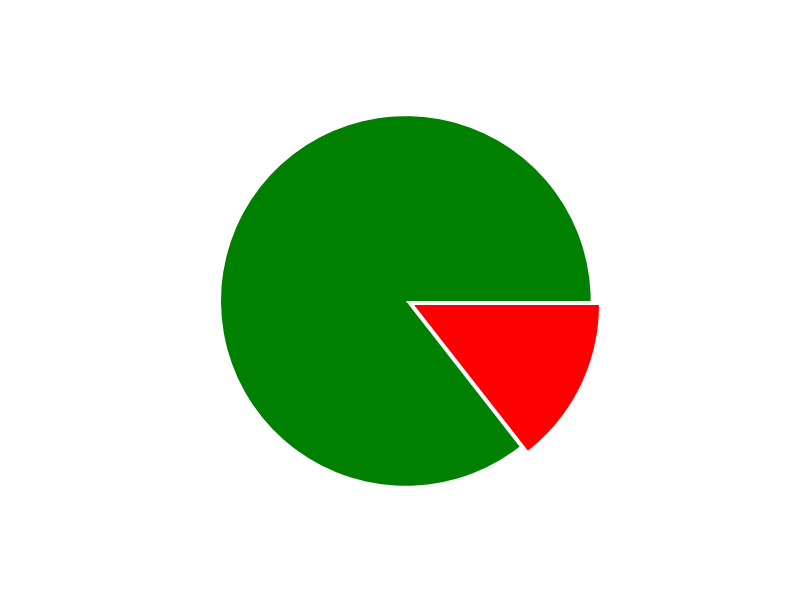}};
        \node (robot_pie) at (7.4,7.7) {\includegraphics[scale=0.42]{robot-imageDeepLabV3plus.png}};
        \node (female_prc) at (2.5,0.5) [align=center] {\small $ \color{Green} \mathrm{TP} = 85.97 \%  ~~~~~ \color{Red} \mathrm{FP} = 14.03 \%$\\ total: 15,607,196 pixels};
        \node (robot_prc) at (7.5,5.5) [align=center] {\small $ \color{Green} \mathrm{TP} = 88.18 \%  ~~~~~ \color{Red} \mathrm{FP} = 11.82 \%$\\ total: 14,873,059 pixels};
        \node (male_prc) at (12.5,10.5) [align=center] {\small $ \color{Green} \mathrm{TP} = 85.55 \%  ~~~~~ \color{Red} \mathrm{FP} = 14.45 \%$\\ total: 15,731,106 pixels};
        \node (label_robot) at (-0.4,7.4) [rotate=90, align=center] {\includegraphics[height=8px]{C1.png} robot};
        \node (label_male) at (-0.4,12.4) [rotate=90, align=center] {\includegraphics[height=8px]{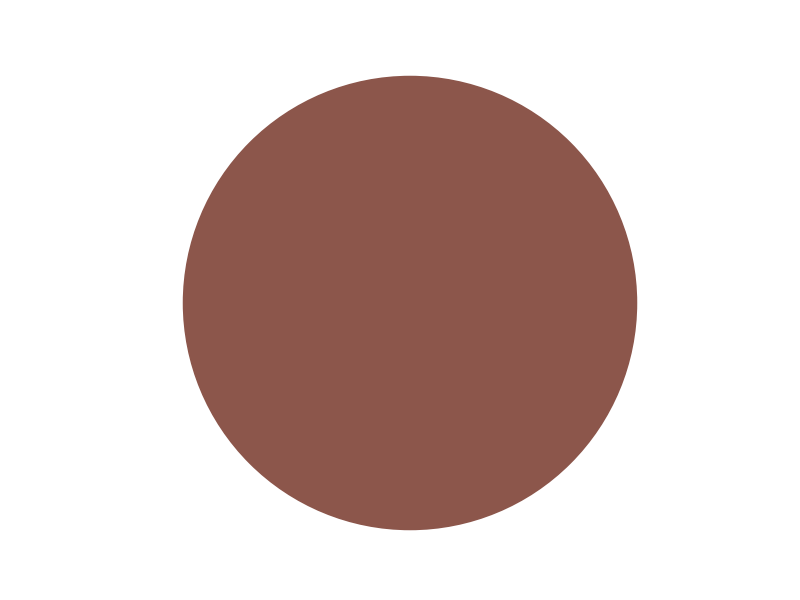} male};
        \node (label_female) at (-0.4,2.4) [rotate=90, align=center] {\includegraphics[height=8px]{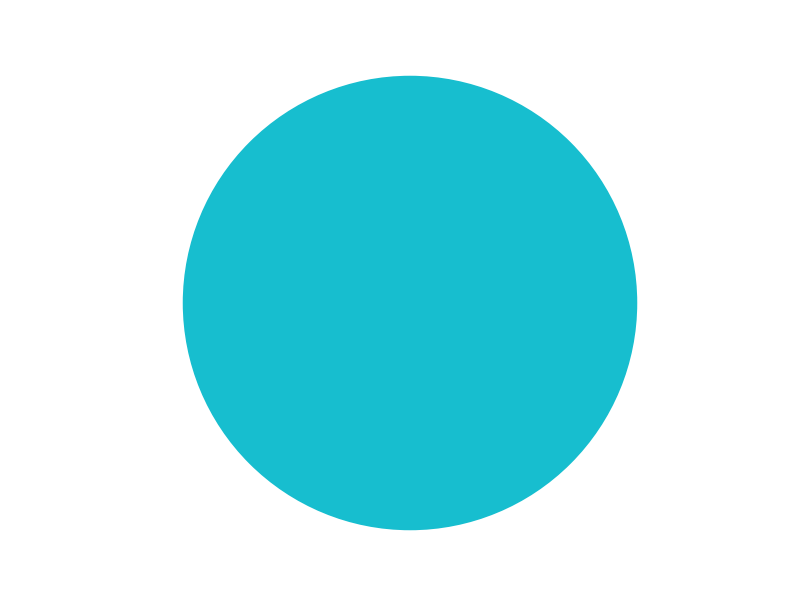} female};
        \node (label_robot) at (7.4,15.4) [align=center] {\includegraphics[height=8px]{C1.png} robot};
        \node (label_male) at (12.4,15.4) [align=center] {\includegraphics[height=8px]{C5.png} male};
        \node (label_female) at (2.4,15.4) [align=center] {\includegraphics[height=8px]{C9.png} female};
        \node [car top,draw=black!,fill=black!90,minimum width=.8cm,rotate=90] at (2.485,10.5) {};
        \node [car top,draw=black!,fill=black!90,minimum width=.8cm,rotate=90] at (7.485,10.5) {}; 
        \node [car top,draw=black!,fill=black!90,minimum width=.8cm,rotate=90] at (2.485,5.5) {}; 
    \end{tikzpicture}
    \caption{Consequential comparisons between the cost assignments of the groups \emph{female}, \emph{male}, and \emph{robot} with respect to the detection of humans. On the diagonal, the pie charts indicate the precision of human predictions as practicability measure for the respective cost assignments. The graphs on the off diagonal display the perceivable area ahead of the ego-vehicle from a bird’s eye perspective. Here, gray crosses show human instances that are overlooked by both considered groups in the comparison, whereas colored dots show human instances that are detected only by the group corresponding to the respective color.}
    \label{fig:female-male-robot}
\end{figure*}

For the consequential evaluation of different cost matrices, we use the safety-aware evaluation tool for the perception of scenes introduced in \cite{brueggemann2021}. We again test the confusion cost matrices from the survey data on the Cityscapes dataset \cite{cityscapes}, which provides image data with humans and their associated distance to the self-driving car. Furthermore, we compare two groups of survey participants against each other as well as the single groups against the robotistic cost valuation / Bayes decision rule.
The evaluation tool generates the bird's-eye view graphics displayed in \Cref{fig:passenger-external-robot} and \Cref{fig:female-male-robot}, in which the complete area in front of the self-driving car as captured by the camera is shown. The described area is of circular form with a center angle of 60°, which is naturally given by the field of vision of the camera.
We subdivide the area in front of the self-driving car into two safety-critical zones, that are 1) the braking distance at a speed of up to 50 km/h and 2) the braking distance at a speed of up to 30 km/h. Accordingly, we assume braking distances of 46.5 meters and 20.6 meters, respectively, which are indicated by the shaded zones in the bird's-eye view graphics.

Moreover, the crosses and dots in the graphics indicate the position of actual human instances relative to the self-driving car.
In this regard, we consider a human instance as {detected} if the associated recall exceeds a threshold of 50\%, \cf \Cref{fig:rec-prc}, otherwise we consider that human instance to be {overlooked}.%
\footnote{\textbf{Disclaimer}: Note that the outlined score is also commonly referred to as detection threshold and it is commonly set to 50\% in the field of computer vision. However, using any other detection threshold between 0\% and 100\% is obviously also possible.}
Then, gray crosses represent human instances that are overlooked by both considered groups in the comparison, while colored dots represent human instances that are detected by only one group.

As discussed in \Cref{sec:Tradeoffs}, the interaction between recall and precision is always an interaction between two opposing metrics.
Therefore, we additionally consider the precision in order to examine to what extent this metric is sacrificed to improve recall. In \Cref{fig:passenger-external-robot} and \Cref{fig:female-male-robot}, the precision is illustrated as pie charts, which we also refer to as \emph{practicability measure} of the self-driving car. In this light, a low precision could yield an autonomous vehicle that mistakenly brakes all the time, since the underlying AI incorrectly identifies human instances all the time, and thus resulting in an impractical application of AI in automated driving.
Via these just described visualizations, we aim at presenting different safety related consequences of AI perception with respect to human classification in a comprehensible and compact way. Besides, we report numerical results in \Cref{tab:results-consequences}.

We have seen in \Cref{sec:results-segmentation} that different confusion cost valuations result in only marginal visual differences in the semantic segmentation masks. However, the consequential evaluation in this subsections reveals that the different confusion cost valuations from the survey yield sufficient changes in AI perception such that the number of overlooked human instances in safety-critical distances to the self-driving car may differ. We observe that the groups \emph{passenger of the self-driving car} and \emph{male survey participants} provide more human sensitive predictions compared to the groups \emph{external traffic participant} and \emph{female survey participants}. The number of overlooked human instances differ by up to 73 within the braking distance of 50 km/h and up to 5 within the braking distance of 30 km/h. More significantly, all the examined groups produce more human sensitive predictions than the \emph{robotistic cost valuation / Bayes decision rule}. With respect to the group \emph{passenger of the self-driving car}, the standard Bayes decision rule overlooks 211 and 13 human instances within the braking distance of 50 km/h and 30 km/h, respectively. As already seen in \Cref{tab:seg-performance}, the reduction of overlooked human instances, \ie gain in recall, is accompanied by a sacrifice of 3.2 percent points in precision. The latter loss could still be considered ``acceptable`', particularly in light of preventing life-threatening street scenarios. 

\begin{table}[]
    \centering
    \begin{tabular}{l||c|ll|ll|l}
    & \multicolumn{1}{c|}{total number of} & \multicolumn{5}{c}{number of overlooked human instances} \\ \cline{3-7}\cline{3-7}
    safety-critical zones & \multicolumn{1}{c|}{human instances} & {passenger} & {external} & {female} & {male} & {robot / Bayes} \\ \hline\hline
    within braking distance of 50 km/h & 2394 & \textbf{534} (13.5\%) & 607 (15.4\%) & 599 (15.2\%) & 550 (13.9\%) & 745 (18.9\%) \\
    within braking distance of 30 km/h & 817 & \textbf{77} (9.4\%) & 82 (10.0\%) & 81 (9.9\%) & 79 (9.7\%) & 90 (11.0\%) \\
    \end{tabular}
    \caption{The amount of overlooked human instances in the Cityscapes validation dataset when using the confusion cost valuations provided by the different groups of survey participants as well as the Bayes decision rule. }
    \label{tab:results-consequences}
\end{table}

\section{Reflection and Limitation of the Survey from a Psychological Point of View} \label{sec:hmi}
In the development of responsible AI technology, active participation and involvement of users and stakeholders take on a new significance. This is especially true if the AI technology is expected to make decisions or act in accordance with human values and not purely based on probabilities, as it is the case in automated driving \cite{dignum2019}, \cf also \Cref{sec:decisionRules-math}. Then, it is not solely a matter of achieving acceptance of the new technology among human users but also a matter of AI technology learning from users, their values and ways of thinking. With the survey conducted in the present study, we aim at gaining insights for the development of AI perception in the field of automated driving by exploring human judgments concerning the costs of potential confusions between object classes.

In traffic psychology, surveys are an established method for uncovering humans’ insights, \eg into their evaluation and usage strategies of driver assistance systems \cite{sprenger2008}. Besides the obvious advantages of this method, self-reports also bear some limitations that must be considered when interpreting the results. In the following, we critically reflect on possible limitations of our conducted survey on different but strongly interacting dimensions (survey and person) in light of the survey participants’ feedback. To this end, we asked participants at the end of the survey to rate the degree of perceived difficulty in completing it (“How complicated did you find the survey?”, 1 = very simple, 2 = fairly easy, 3 = challenging, 4 = complicated, 5 = extremely complicated). On average, participants experienced the survey as challenging ($\textit{mean} = 3.04,\ \textit{standard deviation} = 0.88,\ \textit{number of participants} = 412$). In addition, participants had the opportunity to leave feedback on the survey as free text. The following reflection on the survey builds upon these responses (which we have translated from German into English).

\subsection{Survey-related Factors Influencing the Response Behavior of Survey Participants}

The subject of our survey was particularly complex and challenging. Even though automated driving is an issue of high current relevance, it still appears as a black box for most people and is surrounded by uncertainty. This uncertainty is associated with every innovation and results from a lack of knowledge about the functionality and use of the new technology, see \eg \cite{weick1995sensemaking}. For AI technology, this is especially true because its underlying highly complex algorithms make it impossible to entirely see through and comprehend the AI’s decision making processes, even for experts, see \eg \cite{hagras2018xai}. Due to the limited capacity of human working memory \cite{baddeley1974memory}, at any point in time, only a certain amount of information can be processed simultaneously, so that some information decays. The intransparency of how an AI operates, however, counteracts the development of trust in the technology. Particularly in the field of automated driving, trust in AI’s decision making is vital for acceptance and adoption of this technology in society, see \eg \cite{choi2015trust}.
Accordingly, there is an urgent need for increasing transparency of the underlying models in automated vehicles and, consequently, moving towards an explainable AI \cite{samekxai}. Yet, increased transparency ought not to come at the expense of AI’s decision accuracy. To ensure a high level of decision accuracy, AI technology inevitably requires the processing of large amounts of data far beyond the scope of human comprehension. Thus, there is a fundamental tension between the functioning of the AI being based on a vast amount of data and sophisticated algorithmic models on the one hand, and, on the other hand, the constraints of human cognition, which is usually not able to decode such complex data and models. This causes a dilemma when asking people to partly empathize with an AI, as it was the case in our survey. When designing our survey, we had to find a balance between addressing participants in a way they could still comprehend and, at the same time, not obscuring, but sufficiently capturing the complexity inherent in the topic to obtain meaningful input to feed back into the AI. Thus, despite our best efforts, the survey inevitably involved a certain degree of complexity that might have caused a high cognitive intrinsic load \cite{sweller1994}. Specifically, participants in our study had to make responsible decisions when assessing the level of fatality of confusion between objects in the given traffic scenarios. For every scenario, they had to consider a vast number of aspects and possible consequences for each of the five confusion cases. Hence, participants in our study may not have been able to recognize and process all relevant information sufficiently at a given point in time. Some participants reported this high level of complexity along with substantial uncertainty and problematized it as part of their feedback.

\begin{leftbar}{gray}
\begin{displayquote}
``There are many aspects that must be taken into account. It would be particularly fatal if a person is not recognized as a person and therefore the car does not swerve or brake. With different static or dynamic objects a confusion is less serious in my opinion, because the car should be programmed to swerve in any case. Nevertheless, the issue is very difficult, because it is not possible to consider every aspect and the topic itself is very complex. It is also likely that in retrospect you will notice criteria that you spontaneously did not consider when answering the questions.''
\end{displayquote}
\end{leftbar}

\begin{leftbar}{gray}
\begin{displayquote}
``I struggled to empathize with an AI. What information does it gain from dynamic objects? [...]''
\end{displayquote}
\end{leftbar}

\begin{leftbar}{gray}
\begin{displayquote}
``too many details go into a decision [...]''
\end{displayquote}
\end{leftbar}

Besides the challenging task and content, the special design and technical features of our survey may have caused an increased extraneous cognitive load that is mental load produced by the layout and design of the material \cite{sweller1994}. For example, some participants reported that they had issues determining the weighting of fatality using the sliders and unfamiliar units of value (1-1000000 times more fatal than the confusion car with bus). We chose this scale as it is coherent with the specific design of the used AI model and explicitly intended \emph{not} concealing the actually used confusion costs, like it is the case in the standard Bayes decision rule, \cf \Cref{eq:cost-bayes}.

\begin{leftbar}{gray}
\begin{displayquote}
``For me, it was difficult to determine exactly which slider how to move so that it matched in relation to the other classes. [...]''
\end{displayquote}
\end{leftbar}

\begin{leftbar}{gray}
\begin{displayquote}
``School grades or grading from 1-10 would be easier. As a non-specialist it is very complicated to understand the evaluation criteria...''
\end{displayquote}
\end{leftbar}

\begin{leftbar}{gray}
\begin{displayquote}
``The reference values are difficult to assess and the wording of questions is rather complicated''
\end{displayquote}
\end{leftbar}

To sum up, especially due to the complex characteristics of the topic and task, as well as some design features of our survey, some participants presumably experienced increased intrinsic and extrinsic cognitive load while completing our survey. Consequently, if extraneous and intrinsic cognitive loads were too high, this may have exceeded the limited working memory capacity, so that sufficient cognitive resources that would have been necessary to master the task may not have been available \cite{sweller1994}. In other words, some participants may have been more concerned with understanding the complex scenario and task as well as with becoming familiar with the interface and features of the survey, than with collecting, combining and, weighting relevant information to finally make a sound decision of how fatal they judge the confusion between classes. As a consequence, to save cognitive resources, participants who experienced such cognitive overload may have used less elaborate judgment heuristics, which could risk resulting in poor decision making \cite{wang2007}.

The extent to which the potentially challenging aforementioned features of the survey may have affected participants’ responses might be mediated by personal characteristics. 

\subsection{Personal Characteristics influencing the Response Behavior of Survey Participants}
Various personal characteristics could have determined to what extent the specific features of the survey might have affected participants’ response behavior. For instance, the degree of intrinsic cognitive load essentially depends on whether participants were already familiar with the topics of automated driving or AI in general, and whether they could have drawn on theoretical or even practical experiences. If this has been the case, participants will more likely have perceived the survey as less complex and with less uncertainty due to their existing knowledge than participants who were entirely new to the topic. This is also evident from some feedback reports, in which participants with little or no prior knowledge described their difficulties while completing the questionnaire.

\begin{leftbar}{gray}
\begin{displayquote}
``Despite a short introduction to the topic, I think it is too complicated for non-experts. A simplification [...] might be necessary.''
\end{displayquote}
\end{leftbar}

\begin{leftbar}{gray}
\begin{displayquote}
``Perspectives very technical, more practical explanation for people who really have no idea at all (like me)''
\end{displayquote}
\end{leftbar}

Accordingly, we assume that the more experience and knowledge someone had about AI and automated driving in general, as well as about scenarios like those presented in our survey more specifically, the easier it was to cope with the decision problem of evaluating the fatality of confusions. Participants with relevant experience and prior knowledge likely required less cognitive resources to understand the task and become familiar with the scenarios, allowing them to devote more cognitive resources to the actual decision making process. This would also imply that there might have been some kind of training effect over the course of evaluating the scenarios while completing our survey. This assumption stands in line with some of the participants’ feedback responses.

\begin{leftbar}{gray}
\begin{displayquote}
``You don't understand some questions/situations until you answer the same question several times.''
\end{displayquote}
\end{leftbar}

\begin{leftbar}{gray}
\begin{displayquote}
``Only in the course of the survey did I get a sense of what I would rate stronger or weaker, because I couldn't think much about the consequences beforehand.''
\end{displayquote}
\end{leftbar}

Furthermore, the perceived ease of use of the web-based survey with its technical features probably differed between participants depending on their digital literacy. To be more specific, we expect participants with pronounced digital literacy to have been less distracted and overwhelmed by the technical features of the survey than participants with low digital literacy. Relying on the Technology Acceptance Model, perceived ease of use is one of the two key factors (besides perceived usefulness), which shapes an individual’s attitude towards a technology and therefore serves as a predictor for its adoption \cite{Venkatesh2008TechnologyAM}. Accordingly, perceived ease of use, depending on the degree of digital literacy, could have determined the participants’ attitude towards our survey and consequently their level of engagement. The extent to which the subjects were engaged in their participation in our study certainly also depended on their general interest in topics such as AI and automated driving. In terms of a positive selection, it should be taken into account that the participants who participated voluntarily in our study may have had a general interest in the topic. In turn, interest in and attitudes toward the special topics of AI and automated driving vary with other factors. For example,\cite{Lee2017AgeDI} pointed out, people’s interests for and attitudes toward self-driving cars significantly differ between ages. In their study, they found that older people have, for example, less interest and less confidence in automated driving than young people. Furthermore, gender could have had an influence on both, the attitude towards the tech-centric topic and interface of our survey as well as on the participants’ engagement and response behavior. For example, women, compared to men, tend to have higher computer anxiety and lower computer self-efficacy \cite{Venkatesh2008TechnologyAM}

The list of factors that may have influenced participants in completing the questionnaire is long and cannot be discussed exhaustively in this paper. However, it should have become clear that both the specific nature of our survey and personal characteristics not analyzed in our study may have significantly influenced the participants’ judgments of the fatality of confusion between objects of different classes. Thus, the results of our study should be interpreted in light of these limitations. Nevertheless, our study provides significant insights for the integration of human ethical judgments in the development of AI technology for automated driving and offers several starting points for further research in this field.

\section{The Confusion Cost Matrix from an Ethical Point of View} \label{sec:ethics}

The \Cref{eq:exp-cost-sum} in \Cref{sec:decisionRules} provides the tradeoff space within which the Bayes decision rule amounts to a specific choice with moral significance. Its choice reflects the wish to maximize accuracy in the prediction process, \cf \Cref{eq:cost-bayes}. A choice which prima facie seems to stand in conflict with our moral intuition, which \eg would not set the cost of confusion between a human and a traffic light to be symmetric, as it is the case in the Bayes decision rule according to \Cref{{eq:simple-symm-cost}}. But if the cost of confusion should be distinct from the choice in the Bayes decision rule, what should it be? What would justify a decision rule that would decrease predictive accuracy? Any specific choice will amount to a decision about possible accidents happening or not happening. A feature that it shares with the infamous trolley problem \cite{Foot1967-FOOTPO-2}. 

We will now briefly consider the role of normative theories of ethics and criticisms set forth against the trolley problem and whether they similarly apply to the the choice of a confusion cost matrix. This will be followed by an application of ethical guidelines to the definition of a confusion cost matrix. Finally, we will consider the role surveys may play within an ethical analysis of the confusion cost matrix.

\subsection{Normative Ethics and the Trolley Problem} \label{sec:normative}

Note that the tradeoff space set up by the confusion cost matrix in \Cref{eq:exp-cost-sum} is where a decision has to be made and where a conflict arises between a symmetric confusion cost matrix with our moral intuition.  Normative theories of ethics provide various frameworks within which these choices can be ethically assessed. There are inter alia ethical theories relying on virtues, there are rule-based or deontological accounts and also normative theories that determine the ethically `right' choice by relying on the consequences only \cite{allen2005artificial, wallach2008moral, vallor2016technology}. However, since there is no agreement among ethicists about the ``right'' normative approach (see \cite[p. 676]{himmelreich2018never}) the value of these accounts in individual cases remains contested. 
 
The ethical analysis of the confusion cost matrix relies on two relevant assessment stages, which may or may not impact the moral deliberation. First, the choice of the  confusion cost matrix itself, which, as indicated above, may amount to a morally significant choice with probabilistic consequences (stage 1 assessment). And second, the analysis of what any individual choice of the  confusion cost matrix would actually amount to in concrete cases in terms of possible harms in non-probabilistic terms (stage 2 assessment). One difficulty may arise due to \eg possible assessments of stage 1 choices on the basis of \eg an overall reduction of harm of individuals,  which then may be in conflict with individual stage 2 assessments, which may be concerned with the offsetting of victims -- an ethically problematic procedure. 
The assessment of stage 2 is crucial for a consequentialist analysis of stage 1, while for other non-consequentialist normative theories the stages can in  be assessed independently. Note that the assessment of stage 2 can only play a role in an ex post re-evaluation of the stage 1 choice of the confusion cost matrix. It is the stage 2 assessment, however, which shares features with the trolley problem. We therefore consider now briefly the trolley problem and discuss why issues set forth against its relevance, do not affect our current ethical analysis.
 
In brief, the traditional trolley case is concerned with a trolley that is heading directly towards five people that would die on impact. A lever would allow to switch the tracks of the trolley, such that it would head towards one person and kill that person instead. Should the lever be switched or not? The conceptualization of distinct conflicting moral intuitions regarding what the answer should be has been a much-discussed topic in moral philosophy \cite{foot1967problem,thomson1976killing,kamm2015trolley}. However, these often abstract discussions are of more practical relevance in the context of autonomous vehicles, where these choices may need to be programmed in a certain way. This has led to an extensive debate about the right way to program the autonomous vehicle (see \eg \cite{lin2016ethics,leben2017rawlsian,lin2020robot}). However, recently the practical relevance of the trolley problem has been challenged in \cite{goodall2016away,nyholm2016ethics,himmelreich2018never}.

Following Keeling \cite{keeling2020trolley} we consider two objections to the relevance of trolley cases for autonomous vehicles and whether they would similarly apply to the determination of a confusion cost matrix. The first objection puts doubt on whether the autonomous vehicle will actually encounter trolley cases and so objects to the relevance of moral deliberation for practical applications. Trolley cases like the one mentioned above will amount to extremely rare circumstances and will therefore not be of practical relevance or so the argument goes. Irrespective of whether this is the case for trolley cases, this objection does not apply to the above outlined choice of the confusion cost matrix, as any specific choice of decision rule is a precondition for the application of the algorithm in the first place. As such the decision sets in at an earlier stage in the process of development. Nevertheless, there is a possible analogous objection. Namely, whether the once reasonably restricted choices for the costs of confusions will in effect amount to statistically significant differences in the outcomes. 

To assess this point we can now refer to the results of the survey in \Cref{sec:evaluationMethodology} and the results of the consequentialist analysis in \Cref{sec:results-consequences}. The survey points to differences in the choice of the cost of confusion between gender (male, female) and perspective (passenger, external). Nevertheless, the corresponding segmentation masks only display marginal differences. So even though there are differences in opinions regarding the decision rules, these will not, at least prima facie, significantly change what the AI can ``see''. However, this prima facie irrelevance of the survey-based confusion matrix does not remain irrelevant once one considers the full consequentialist assessment of these decision rules. Here it becomes apparent that the survey-based decision rules do, unsurprisingly, a better job in recognizing humans compared to the Bayes decision rule. So an analysis of possible variations of the decision rule do lead to significant differences of relevance for moral deliberation. Note that even if there would not have been a difference, the irrelevance objection would not apply, as for the determination of the decision rule, the irrelevance claim is a result of the assessment and not something that can be applied beforehand. 

The second objection to trolley cases is the moral difference argument according to which the moral deliberation regarding the choice between pushing the lever or not, differs in a morally significant way from the deliberation in real world cases. In real world cases there are additional circumstances that would impact the moral deliberation of the tradeoff space. For instance  the manufacturers may have certain obligations (\eg protecting the safety of the passengers) or usually  the outcomes are not known a priori, but may only be associated with probabilities. As has been argued in \cite{nyholm2016ethics,himmelreich2018never}, reasoning under uncertainty differs significantly from reasoning about known facts. This is a significant objection to trolley cases that similarly impacts the deliberation on the possible choices in our tradeoff space. 

One may abstractly morally deliberate on certain choices of the confusion cost matrix, but these would be significantly impacted by additional information about the possible outcomes that would be associated with these choices. While we did consider the consequences of specific choices of the confusion cost matrix in \Cref{sec:results-consequences}, thereby mitigating this objection to some extent, it remains a possibly categorical objection to the significance of survey results, where limitations to the human capability to reason with probabilities and uncertainties will restrict the possible value of these results. This is nicely illustrated by the somewhat surprising result of the consequentialist analysis that assigning generically high costs in the survey has a mitigating effect in valuing human lives. A consequence that probably was not foreseen by the survey participants aiming at a more conservative driving attitude.

The moral difference argument therefore points to the possibly problematic nature of the already discussed difference between the stage 1 and 2 assessments. However, the consequentialist analysis in \Cref{sec:results-consequences} provides a tool that does address this problem head-on by making transparent the possible moral differences in the choices in the survey compared to an assessment of their corresponding consequences. A lesson one may learn from this, is that for more reliable survey results, \ie survey results which more completely capture the individuals' choices, one may need to complement the determination of the cost confusion matrix with the respective birds-eye view pictures of the corresponding consequences.

 \subsection{Ethics Guidelines of the Ethics Commission on Automated and Connected Driving} \label{sec:ethical_guidelines}
 
In 2016 an ethics commission on automated and connected driving was appointed by the German federal minister of transport and digital infrastructure to develop ethical guidelines for automated and connected driving. In June 2017 they published a report with 20 ``ethical rules for automated and connected vehicular traffic'' \cite{difabio2017ethics}. Since then many more guidelines have appeared, significantly in the European context the recommendations of the ``Ethics of connected and automated vehicles’’ report by an independent expert group implemented by the European commission (EC) \cite{bonnefon2020ethics}. We will now consider the relevant rules and what they imply for the choice of the confusion cost matrix and the inclusion of a survey-based approach.

Ethical guidelines provide recommendations that are often intentionally stated at an abstract level to ensure generality, while at the same time need to be concrete enough to lead to actionable advice. This presents a tension that is not always easy to implement. For instance, rule 5 of the German report states that the ``[a]utomated and connected technology should prevent accidents wherever this is practically possible'', and the design and programming of the vehicles ``be such that they drive in a defensive and anticipatory manner, posing as little risk as possible to vulnerable road users''. One difficulty in implementing this rule is the ambiguousness regarding what may be considered ``practically possible'' or what ``as little risk as possible'' mean. Both terms do not recognize the tradeoff situation present in these situations. The term ``as little risk as possible'' may suggest that it is a problem of minimization, where in fact it is an optimization problem, where various values and their respective moral considerations need to be balanced.

Being aware of this predicament, the report of the EC expert group recognizes, after listing the guiding ethical principles, that ``[t]he above principles cannot be applied with a mechanical top-down procedure. They need to be specified, discussed and redefined in-context.’’. They further note that this ``is why the design and development of CAV (connected and automated vehicle) systems should be supportive of and resulting from inclusive deliberation processes involving relevant stakeholders and the wider public.’’. That is, the concretization and implementation of any individual recommendation should not be the decision of individual scientists but include the ``wider public'' (as it is aimed with the survey-based approach in \Cref{sec:survey} and \Cref{sec:evaluationMethodology}) and other stakeholders (see industry perspective in \Cref{sec:industry}) as well as being ``supportive of’’ the inclusive deliberation process (which we implement by considering the limitations and capabilities of the survey-based approach in \Cref{sec:hmi}). 
 
The starting point of our analysis is in line with recommendation 14 of the EC expert group, which requires a reduction of opacity in algorithmic decisions, a recommendation necessitated by the fact that ``algorithm-based CAV systems [...] may operate as ``black-boxes'' that do not allow cognitive access to how they have arrived at a particular output, or what input factors or a combination of input factors have contributed to the decision-making process or outcome'' \cite[p. 49]{bonnefon2020ethics}. By recognizing the implicit tradeoff space within which the Bayes decision rule is making a morally significant choice one opens up the ``black-box'' to scrutiny. This has led us to the two assessment stages, which in turn now need to be confronted with what the other ethical rules and recommendations imply for them. 

So let us now turn to the relevant rules and recommendations of the two guidelines regarding the assessment of the two stages. Rule 7 of the German report states that ``within the constraints of what is technologically feasible, the systems must be programmed to accept damage to animals or property in a conflict if this means that personal injury can be prevented''. This statement addresses the stage 2 decision regarding the offsetting of various scenarios. This may be translatable, and therefore technologically feasible by setting strong asymmetric cost functions, valuing the human category significantly higher than other entities. For that one needs to consider the relation between choices of the confusion cost matrix and their consequences in specific scenarios. Something already considered in \Cref{sec:results-consequences}. This, however, remains in conflict with the general practicability of the autonomous vehicle (AV), \cf \Cref{sec:Tradeoffs}, and thus needs to be put in balance with a justification for the practicability of the AV. An impracticable AV would not be introduced and therefore could not provide ``promises to produce at least a diminution in harm compared with human driving, in other words a positive balance of risks'' (rule 2). This is recognized as a necessary part of the justification of AVs in general. 
 
In Rule 9 the ethics commission formulates a clear prohibition ``to offset victims against one another''. They provide two reasons for that. First, there is a simple practical issue mentioned in rule 8: ``a decision between one human life and another, depend on the actual specific situation, incorporating “unpredictable” behaviour by parties affected. They can thus not be clearly standardized, nor can they be programmed such that they are ethically unquestionable''. This practical reason is complemented with an underlying Kantian argument speaking against a pre-programmed stage 2 decision. From a Kantian perspective an individual enjoys the right of moral self-determination, a right that the person would be robbed off in case of an externally determined decision, which ``in extremis, [would] be able to take correct ethical decisions on the demise of the individual human being''. Rule 9 therefore amounts to a prohibition to base the confusion cost matrix on the basis of individual stage 2 decisions. 
 
The ethics commission, however, continues in rule 9 by stating that a ``[g]eneral programming to reduce the number of personal injuries may be justifiable''. The reason for that is that a general programming of this kind would be abstract in the sense that it does not take the life of any specific individual into consideration, as these are not known at this level, and further ``the programming reduce[s] the risk to every single road user in equal measure'' \cite[p. 18]{difabio2017ethics}. As such a stage 1 decision that would only be statistically determined by individual stage 2 assessments would be in line with the recommendations of the ethics commission, as long as overall personal injuries would be reduced.

The report by the EC expert group takes a more general approach regarding dilemma situations, by considering acceptable behaviour in dilemma situations as something that can ``organically emerge from the adherence to the principles of risk distribution stated in Recommendation 5''. Risk-distribution in recommendation 5 allows for a differential behaviour around a certain subset of road users in cases where these belong to certain vulnerable groups. A different behaviour of the CAV regarding these vulnerable groups then amounts to an opportunity to redress inequalities present in current traffic collisions. That is, it would allow one to adequately account for the ethical issues involved with the consideration of \eg wheelchair users, cyclists and visually impaired users. This is of course only a viable option, if the algorithm has the capability to distinguish between these groups. In the case of the cost confusion matrix above, the corresponding classes do not distinguish between possible vulnerable groups except the distinction between the classes ``human'' and ``dynamic''. At this stage the report again refers to the necessity to discuss the ethical and social acceptability ``as a topic for inclusive deliberation'' \cite[p. 31]{bonnefon2020ethics}. This now leads us to the survey-based approach and its ethical ramifications.

\subsection{Surveys, the Moral Machine Experiment, and Ethics}\label{sec.survey_ethics}

The moral machine experiment is a massive survey-based analysis of the trolley problem \cite{awad2018moral}. By including more than 40 million decisions by over two million participants from over 200 countries it provides an extensive analysis about the moral intuition of people around the world regarding the trolley problem and its variations.
It has since generated a lot of discussion and criticism, largely due to the framing of the trolley setups. Providing scenarios where one may have to choose between athletes \vs overweight persons or executives \vs homeless persons may lead to information about the corresponding people and their cultures but less on what the ethics of the AI should be, as the introduction of these distinctions in the classification process could already be in violation of basic human rights (see \eg \cite{bigman2020life,kochupillai2020programming,jaques2019moral, furey2021s} for this and other objections against the moral machine experiment). 

Even in less problematic decision processes (\eg decisions in dilemma setups involving human \vs non-human entities, which would be in agreement with rule 7 of the German report) there are limitations to drawing conclusions from the moral intuition of individuals to what ought to be the case. This is true both in the case of more individual implementations of the moral intuition (possibly regionally restricted) as well as the identification of moral intuitions shared by the majority. The former may lead to a moral relativism while the latter may result in the corresponding systematic discrimination of vulnerable minority groups. 

So what role can surveys play in the ethical analysis of tradeoff situations? To consider one of these roles let us briefly introduce another variant of the trolley problem, namely the tunnel problem \cite{ millar2015technology, millar2017ethics}: Imagine being the passenger of an autonomous vehicle approaching a tunnel, where all of a sudden a child mistakenly runs in front of the tunnel. There are two options, either one hits the child or one swerves to either direction colliding with the wall of the tunnel. So either the child dies or the passenger of the vehicle dies. What this problem points to is the very personal nature of this choice. There is again no clear ethical solution to this case and for either decision there might be good reasons to give. 

This has led Millar \cite{millar2017ethics} to suggest an analogy to the case of end-of-life decisions in the healthcare context. Medical professionals are required to seek informed consent from the patients and are not simply allowed to decide for the patient. Similarly, Millar argues, one should incorporate the individuality of the people involved in the decision-making process and it should not be the decision of the engineer or manufacturer of the autonomous vehicle. The EC expert group report does explicitly incorporate in recommendation 8 that one should ``[e]nable user choice, seek informed consent options and develop related best practice industry standards’’, where consent-based user agreements should ``go beyond ``take-it-or-leave-it’’ models of consent.’’ This together with the already mentioned necessity for inclusive deliberation processes involving the wider public introduces an important ethical role to survey-based approaches.

\section{Participatory Approach to Ethics from the Point of View of the Automotive Industry} \label{sec:industry}
One of the major reasons for assessing the quality of perception AI within industrial applications lies with the validation of safety goals. For safety-critical automotive software systems such as advanced driver assistance systems (ADAS), a \textit{positive risk balance} is aimed at, \ie the risk stemming from the usage of a system shall be less than the risk while not using the system. Furthermore, \textit{unreasonable risk} is to be avoided. Here, risk is called \emph{unreasonable}, if it is ``unacceptable in a certain context according to valid societal moral concepts'' \cite{26262}. Hence, driving notions behind arguing safety for ADAS compare the behavior of a system with the respective human behavior. Positive risk balance and avoidance of unreasonable risk are then argued by careful analysis, development of rigor as well as quick update processes in the case of detected safety issues, and extensive testing.

Different standards aiming at particular aspects of \emph{safety} for automotive software and AI therein are being developed, see \eg \cite{4804}, ISO/AWI PAS 8800, and ISO/AWI TS 5083 that are under development (at the time of writing this work).
Next to standardization institutions, also legislation is putting regulatory frameworks into place. One such example is the EU AI act, which is currently in preparation. It involves important guidelines, \eg with respect to the constitution of test sets and AI behavior. This involves bias-freeness, fairness as well as statistical representativity and diversity both in training and validation data. It also advises human oversight and assessment of the performance.

In this light, those outlined standards could represent some form of instructions that software engineers should follow when programming AI software. Therefore, on the one hand, standards need to take ethics guidelines developed by expert groups into consideration as described in \Cref{sec:ethical_guidelines}. On the other hand, developing standards additionally also involves the need to understand the social expectation towards safe system behavior and its derived performance requirements on (AI-driven) perception systems.
The approach examined within this work, \cf \Cref{sec:survey}, presents a possible way to understand the required AI-based perception performance as it allows for understanding social expectations towards the semantic perception interpretation in a democratized way. Thereby, this approach may help to raise public acceptance as well as trust for AI-based systems for automated driving. 

The findings of this work indicate a clear discrepancy in the results when comparing the participatory approach with the robotistic approach. Relevantly in terms of safety, the perception system resulting from the participatory approach clearly detects more safety critical pedestrians than that from the robtistic approach, while keeping the false positive rate at almost the same level. This points towards the presented approach being a possible candidate for the design process of systems for automated driving (possibly even in both stage 1 and stage 2 assessment, \cf \Cref{sec:normative}).

Nonetheless, to to be fully usable from an industrial point of view, some future work is still required:
\begin{itemize}
    \item[\textbf{1)}] Safety \emph{always} relates to the overall system and not to particular sub-components like the perception. It also allows for inner-system resilience, \eg unreasonably erroneous perception results can be mitigated by the trajectory planner. Hence, the desired perception behavior needs to be assessed within the context of the overall system for deriving safety requirements. 
    \item[\textbf{2)}] Clarity with respect to the given survey tasks has to be ensured. Otherwise the interpretations of the survey results might be misleading.
    \item[\textbf{3)}] Alternative evaluation schemes such as comparison to human perception, \eg "What do humans see?" rather than "What do humans expect the robot to see?", could also be taken into consideration to derive an improved understanding of the social expectation to perception performance.
\end{itemize}

\section{Conclusion and outlook} \label{sec:conclusion}

\subsection{Conclusions}
In this work we explored a problem of practical ethics in the context of automated driving, namely the definition of the cost of confusion of different categories, such as humans, dynamic objects, road, and others. As one solution strategy, we experimentally explored a participatory approach, \ie we ask people to enter the values for the cost of confusions in an online survey. We developed statistical tests to prove that different groups of survey participants, \eg by gender or perspective (passenger \vs pedestrian), differ significantly in their cost valuations. Moreover, we show that cost structures of all survey participants as unit differ even more significantly from the ``robotistic'' view, in which all confusions come with the same cost. We demonstrated that the different cost structures change the perception of an system of artificial intelligence (AI), which possibly results in individual human instances, within a safety critical distance relative to the self-driving car, to be poorly detected or even being entirely overlooked. 

We also explored the limitations of the presented participatory approach from three different disciplinary perspectives. We therefore embed our approach in a broader discussion of practical ethics, investigate the psychological difficulties of taking such decisions on the cost structure, and include feedback from the automotive industry.

Whether participatory elements can become an element in an \emph{ethics by design} approach to AI or whether it can be taken into consideration in regulatory frameworks for AI technology depends on the resolution of several issues that were described in this article.

First of all, further investigations are needed to better understand the potential role of participatory elements for the normative restriction of AI design approaches, let it be in the design step itself or in regulatory texts like technical norms. Here, a deeper understanding of the influence on the setup for participation, such as the survey design, is important. In particular, the difficulties for people to take a decision on the configuration of an AI system without the ability to be aware of all potential consequences needs further consideration.

\subsection{Recommendations for Future Research}
The survey conducted in the present study is the result of a weighing process between a design that is understandable for the participants on the one hand and still meets the necessary depth to sufficiently depict the confusion costs for the training of the AI model on the other hand. Maybe future studies can still further refine the questionnaire towards an increased usability without undermining the required level of data complexity. Special attention should then be paid on reducing participants’ extraneous cognitive load as far as possible. For instance, this could be achieved by avoiding unfamiliar technical features such as the sliders. Moreover, in future work visualizations should depict the respective perspective from which the participants are asked to assess scenarios. In the present study, participants assigned to the group of external traffic participants were not given different visualizations, \ie they also view the visualizations from the perspective of car passengers and were only prompted by a textual cue to imagine being outside the car as a pedestrian. This means that in order to evaluate the situation from their assigned role, they need to perform an extra step of mental elaboration, namely switching from the shown perspective to their assigned perspective. This is in contrast to to the participants in the group of car passengers, who viewed the visualizations from their assigned perspective. 

The question whether an improved survey design changes the outcome of the survey merits an investigation.
To fully capture the moral intuition of the involved participants in both the stages of decision -- defining the cost structure and the resulting outcomes, respectively -- should be part of any future survey process, to ensure that the participants are aware of the consequences of individual confusion cost determinations. 
Further, by making explicit the underlying confusion cost structure in decision rules, one introduces the opportunity not only to adapt and incorporate human ethical intuition but also go beyond it and improve it. Future research should consider the possibility to include more differentiated classes in the analysis to allow, following recommendation 5 of the EC expert group, to rectify existing inequalities where vulnerable groups are involved.

Besides the analysis of deviations from the Bayes decision rule, one should also incorporate further values in the analysis. Any choice of the confusion cost matrix based on a participatory approach needs to be systematically balanced with other important values like practicability (see \Cref{sec:results-consequences} for a first step in that direction) and speed. Note that this is not a balancing of ethical decisions with non-ethical ones. As already mentioned, practicability is a precondition for the introduction of self-driving cars in the first place, which in turn can lead to an overall reduction of harm and so is part and parcel of the moral calculus. The moral dimension of these often technical features is not always apparent from the get-go and interdisciplinary work may allow to reveal the moral aspects of these technical features. This kind of moral evaluation of technical features, the systematic comparison of them, and the incorporation of other value-laden features should be considered in future research. 

In the present study, we used static traffic images to compare human with robotic perceptions based on the different confusion judgments. Building on this, future studies should investigate human confusion judgments under real-life conditions in a simulation-based environment, such as \eg presented in \cite{kowol2022aeye}. The traffic scenarios to be evaluated are dynamic in nature, accordingly, a simulation-based environment could help participants to put themselves in the complex situation, to estimate consequences of confusions and to make immediate decisions authentic. We therefore recommend complementing the insights drawn from the self-reported data of this study with behavioral data from a simulation-based environment in future research. 

In general, the understanding and the shaping of relations between normative ethics, technical norms, the public's opinion, and design decisions for AI in safety relevant applications, such as automated driving, pose interesting challenges for the future. This equally applies to research, academic, and political debate.

\vspace{.5cm}

\begin{wrapfigure}[6]{r}{0.3\textwidth}
  \begin{center}
    \vspace{-27pt}
    \href{https://www.bmwi.de/Navigation/EN/Home/home.html}{\includegraphics[height=108pt]{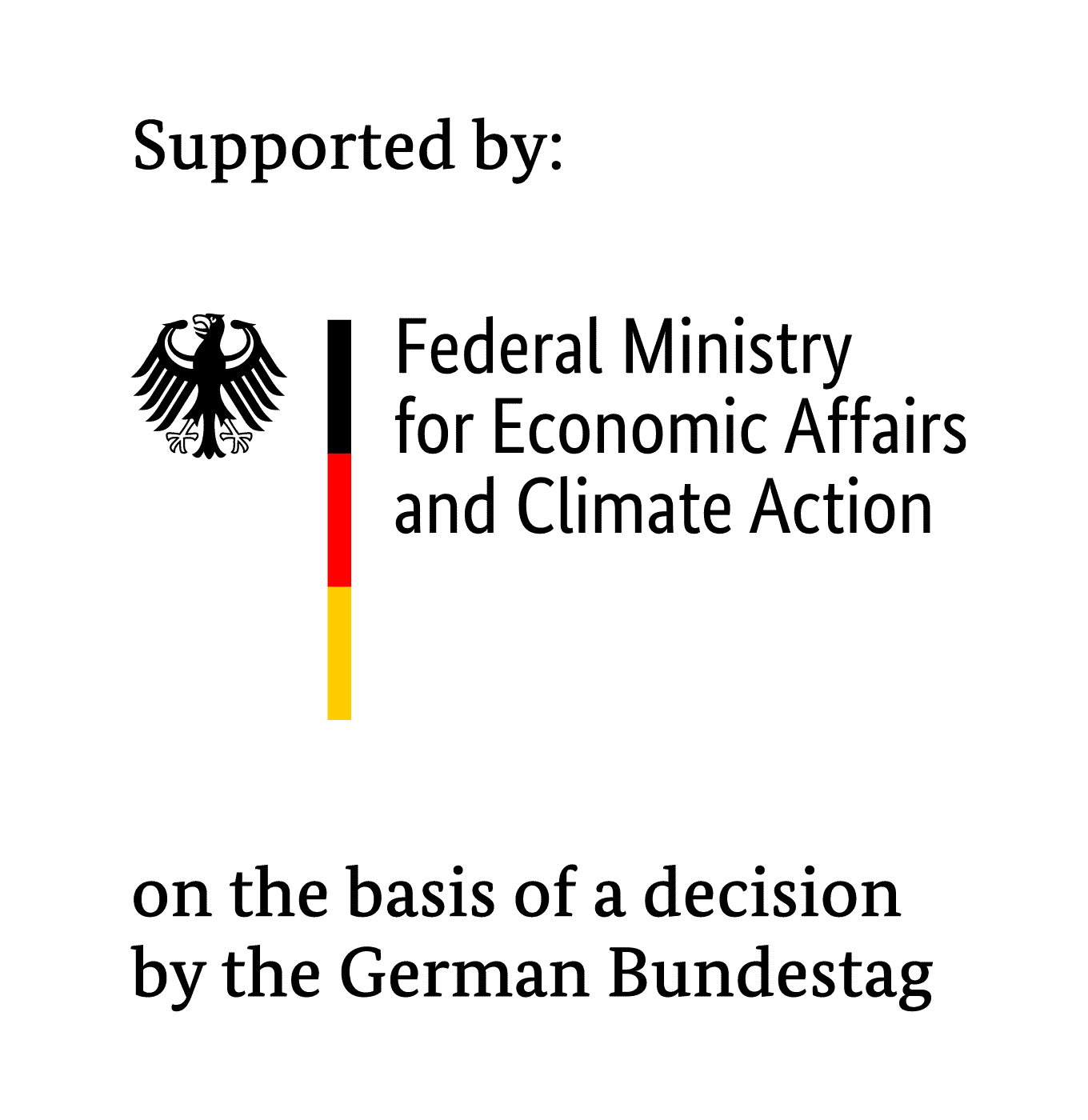}}
  \end{center}
\end{wrapfigure}
\noindent \textbf{Acknowledgement.} Robin Chan, Dominik Brüggemann, Matthias Rottmann, and Hanno Gottschalk acknowledge partial financial support by the German Federal Ministry for Economic Affairs and Climate Action within the project ``Methoden und Maßnahmen zur Absicherung von KI basierten Wahrnehmungsfunktionen für das automatisierte Fahren (KI-Absicherung)", grant no. 19A19005R. The authors would like to thank the consortium for the successful cooperation.

\newpage
\bibliographystyle{IEEEtran}
\bibliography{biblio}

\newpage
\appendix

\subsection{Modified F-test to test for statistical Differences between two Confusion Cost Matrices} \label{app:f-test}
Let $x_{l,k,j,i} \in \{ 0,1,2,3,4,5,6\}$ denote a single entry from the survey data where the indices correspond to the following sets:
\begin{align*}
    l \in \{ 1,2 \} &: \textrm{group index}
    &k \in \{ 1,2,\ldots,6 \}&: \textrm{true target class index} \\
    j \in \{ 1,2,\ldots,6 \}&: \textrm{confused class index}
    &i \in \{ 1,2,\ldots,n_{l,k} \}&: \textrm{provided answer index} ~,
\end{align*}
where $n_{l,k} \in \{ 1,2, \ldots, n \}$ denotes the number of provided answers from group $l$ assessing confusions with target class $k$.
Further, we denote the overall mean entry for the confusion of target class $k$ with class $j$ by
\begin{align}
    \Bar{x}_{\boldsymbol{\cdot},k,j,\boldsymbol{\cdot}} = \frac{1}{n} \sum_{i=1}^{n} x_{\boldsymbol{\cdot},k,j,i}
\end{align}
and the mean entry within group $l$ for the confusion of target class $k$ with class $j$ by
\begin{align} 
    \Bar{x}_{l,k,j,\boldsymbol{\cdot}} = \frac{1}{n_{l,k}} \sum_{i=1}^{n_{l,k}} x_{l,k,j,i} ~.
\end{align}
Hence, the mean sum of squared differences \emph{between} groups is computed via 
\begin{align}
    MS_B &:= \sum_{k=1}^6 \sum_{j=1}^6 ~ n_{1,k}(\Bar{x}_{1,k,j,\boldsymbol{\cdot}} - \Bar{x}_{\boldsymbol{\cdot},k,j,\boldsymbol{\cdot}})^2
    + \sum_{k=1}^6 \sum_{j=1}^6 n_{2,k}(\Bar{x}_{2,k,j,\boldsymbol{\cdot}} - \Bar{x}_{\boldsymbol{\cdot},k,j,\boldsymbol{\cdot}})^2
\end{align}
while the mean sum of squared differences \emph{within} groups is computed via
\begin{align}
    MS_W &:= \sum_{k=1}^6 \sum_{j=1}^6 \sum_{i=1}^n \frac{(x_{1,k,j,i} - \Bar{x}_{1,k,j,\boldsymbol{\cdot}})^2} {\sum_{k=1}^6 n_{1,k} + n_{2,k} - 60} + \sum_{k=1}^6 \sum_{j=1}^6 \sum_{i=1}^n \frac{(x_{2,k,j,i} - \Bar{x}_{2,k,j,\boldsymbol{\cdot}})^2} {\sum_{k=1}^6 n_{1,k} + n_{2,k} - 60}
\end{align}
with 60 being the degrees of freedom (the number of entries in the two $6 \times 6$ confusion cost matrices that are not fixed, \ie the off main diagonal entries).
Then, the $F$-statistic 
\begin{align}\label{eq:f_value}
    F := \frac{MS_B}{MS_W} = \frac{\textit{between-groups variance}}{\textit{within-groups variance}} \in \mathbb{R}_{\geq 0}
\end{align}
is defined as the ratio of \emph{between} groups variances to \emph{within} groups variances.
We apply a bootstrapping method in order to approximate a \emph{p-value}
\begin{equation}
    p = \frac{\#\{ F_{random} > F \}}{\#\{ F_{random} \}} = \frac{\#\{ F_{random} > F \}}{S} \in [0,1]
\end{equation}
with $F_{random} \in \mathbb{R}_{\geq 0}$ the random $F$-statistic (according to \Cref{eq:f_value}) after shuffling the group affiliations of the survey data and $S \in \mathbb{N}$ the number of shuffling steps. The lower the p-value the more likely the investigated groups' means differ.

\end{document}